%% file: main.tex
\documentclass[11pt,a4paper]{article}

\usepackage{authblk}
\usepackage[utf8]{inputenc}
\usepackage[hyperref]{naaclhlt2018}
\usepackage{times}
\usepackage{latexsym}
\usepackage{framed}
\usepackage{url}
\usepackage{todonotes} %[disable]{todonotes}
\usepackage{multirow}
\usepackage{booktabs}
\usepackage{subcaption}

\aclfinalcopy
\def\aclpaperid{504} %  Enter the acl Paper ID here

\author[1]{\textbf{James Thorne}}
\author[1]{\textbf{Andreas Vlachos}}
\affil[1]{Department of Computer Science, University of Sheffield}

\author[2]{\textbf{Christos Christodoulopoulos}}
\author[2]{\textbf{Arpit Mittal}}
\affil[2]{Amazon Research Cambridge}

\affil[ ]{\texttt{\{j.thorne, a.vlachos\}@sheffield.ac.uk}}
\affil[ ]{\texttt{\{chrchrs, mitarpit\}@amazon.co.uk}} 

\DeclareCaptionFont{10pt}{\fontsize{10pt}{12pt}\selectfont}
\date{}

\title{FEVER: a large-scale dataset for Fact Extraction and VERification}

\begin{document}

\maketitle

\begin{abstract}
%Large volumes of useful information are available online from sources with questionable reliability.
%AV: Not sure I like to focus on reliability because many call Wikipedia unreliable
%It is vital to be able to verify the extracted information and flag potential sources of misinformation. 

%\todo[color=yellow]{I suggest removing this 1st sentence}Verification of textual claims, often referred to as fact checking, is the task of assessing the truthfulness of statements in natural language. %, and it considered to be one of the main methods in combating misinformation. 
%AV: I don't mind talking about other applications, just think this is the most catchy one to put in the abstract
%\todo{
%\todo[color=yellow, author=Christos]{I suggest removing the first sentence} Unlike other tasks
%, ?}% such as question answering and textual% entailment, %JT - is this part of the sentence Andreas? not sure if it made sense in the way it was written, especially with the next line starting with a captial letter
%Despite its importance,
%and despite recent interest, 
%research in textual claim verification has been hindered by the lack of large-scale manually annotated datasets.
In this paper we introduce a new publicly available dataset for verification against textual sources, FEVER: Fact Extraction and VERification. It consists of 185,445 claims generated %\todo{I prefer constructed, Sec3 calls it generated} 
by altering sentences extracted from Wikipedia and subsequently verified without knowledge of the sentence they were derived from.
%AV: careful, they knew the page
%in a two-stage process. Initially, claims were constructed by manually altering sentences extracted from Wikipedia, thus obtaining claims not necessarily being supported by the original sentence. Following this, these claims were verified against Wikipedia by annotators who did not know the sentence they were constructed from, 
The claims are classified as \textsc{Supported}, \textsc{Refuted} or \textsc{NotEnoughInfo} by annotators achieving 0.6841 in Fleiss $\kappa$. 
 For the first two classes, the annotators also recorded the sentence(s) forming the necessary evidence for their judgment.
%This is an annotated corpus of 125391 true and false human-generated facts with supporting and refuting evidence from Wikipedia.
%Our dataset is constructed to jointly address open-domain Information Extraction and Fact Checking using either a symbolic or distributional approach to reasoning. AV: Too technical for the abstract I think, let's save it for the main text
%AV: could you hint at what you mean? maybe just give examples of the methods you are referring to
%JT: Done above 
%AV: Let's shorten this
%In this paper we introduce 
%(simple subject-predicate-object triples expressed in natural language) %AV: We are doing more complex claims, actually that's an important departure from previous work
%both of factual information found in Wikipedia, but also \textit{mutated information} (negations, paraphrases, generalizations, ontological substitutions).
%AV: Maybe too much info. Why not focus on the fact it was manually done
%For each verifiable claim (approx three-quarters of the dataset), %AV: Give the numbers, maybe only mention this number.
%AV: First say the three way classification, then mention the evidence for the support/refute
%we provide evidence from Wikipedia, annotated at a sentence level that supports or refutes the claim. We designed-in the controlled use of world knowledge from Wikipedia as part of our claim-generation process: \todo{What percent} XX\% of these claims require compositional knowledge from 2 or more different Wikipedia pages to verify.
To characterize the challenge of the dataset presented, we develop a pipeline approach 
%using both baseline and state-of-the-art components 
%that given a claim retrieves relevant documents, then selects evidence containing sentences, and finally determines the label of the claim w.r.t. Wikipedia. 
and compare it to suitably designed oracles. % and manual analysis we observe that the main challenges are ...\todo{}.
%To characterise the challenges of predicting the truthfulness of information given evidence, %AV: just say challenges of the task
%we have used a constructed a pipelined system that comprises both simple and state-of-the-art information retrieval and textual entailment components. We found that \todo[inline]{What are the main findings?} 
%AV 
%AV: Maybe say what is the hardest thing given the oracle experiments?
The best accuracy we achieve on labeling a claim accompanied by the correct evidence is 31.87\%, while if we ignore the evidence we achieve 50.91\%.
Thus we believe that FEVER is a challenging testbed that will help stimulate progress on claim verification against textual sources. %\todo[color=yellow]{Make this stronger: FEVER is a challenging test-bed}
%, while the insights obtained during the annotation process will be useful to other large-scale annotation efforts.
%We believe that the FEVER corpus will be a good test-bed for knowledge extraction \todo[inline]{any other system} systems of the future.
\end{abstract}

\input{1_intro}
\input{2_related_works}
\input{3_new_corpus}

\input{4_baseline}
\input{5_evaluation}

\input{6_discussion}
\input{7_conclusions}

%\todo{AV: In the manual error analysis, we should highlight that systems getting it right for the wrong reasons is quite common.
%AV: see Ling's Rationale generation paper. Also worth citing Grefenstette's book story challenge
%JT: Described - although without citing.
%}

\begin{footnotesize}
\section*{Acknowledgments}

The work reported was partly conducted while James Thorne was at Amazon Research Cambridge. Andreas Vlachos is supported by the EU H2020 SUMMA project (grant agreement number 688139). The authors would like to thank the following people for their advice and suggestions: David Hardcastle, Marie Hanabusa, Timothy Howd, Neil Lawrence, Benjamin Riedel, Craig Saunders and Iris Spik.  The authors also wish to thank the team of annotators involved in preparing this dataset.

\end{footnotesize}

\bibliography{references}
\bibliographystyle{acl_natbib}

%Commented out to stop my browser slowing down
\clearpage
\appendix
\input{a_appendix}

\end{document}

%% file: 1_intro.tex
%auto-ignore
\section{Introduction}

%AV mention multihop: We designed-in the controlled use of world knowledge from Wikipedia as part of our claim-generation process: \todo{What percent} XX\% of these claims require compositional knowledge from 2 or more different Wikipedia pages to verify.
%AV: orders of mangitude
%AV: pipeline description: that given a claim retrieves relevant documents, then selects evidence containing sentences, and finally determines the label of the claim w.r.t. Wikipedia. 
%AV: Maybe say what is the hardest thing given the oracle experiments?
%AV: while the insights obtained during the annotation process will be useful to other large-scale annotation efforts

The ever-increasing amounts of textual information available combined with the ease in sharing it through the web has increased the demand for verification, also referred to as fact checking. While it has received a lot of attention in the context of journalism, verification is important for other domains, e.g.\ information in scientific publications, product reviews, etc.
%AV: removed this to save space since we don't do anything about automatic IE here.
%, especially since there has been substantial progress in extracting information automatically from such sources.
%AV: I like the one below, but I find it easier to justify what we do from the verification perspective. But it is obviously the case that verification assumes extraction and it should be stressed in the intro. 
%the ability to extract accurate and consistent facts is paramount to ensure that human language technologies such as news aggregation, semantic search and question answering provide correct information to end-users.  As these downstream applications become reliant on extracting information from user-contributed or unverified sources, there exists a necessity to verify the extracted information. 
%AV: could integrate this too if we have space:
%However, both the diversity and volume of available information precludes human moderators from manually verifying all facts that are ingested into a KB.  
%Both characterizing and improving the level of trust in automatically generated structured information will be instrumental in ensuring accurate knowledge is  KBsystems .limiting semantic drift %AV: people use this to refer to different things, what do you mean
%and ensuring high precision %AV: accuracy? (precision can always be improved at recall's expense)
%Knowledge Base Population.

In this paper we focus on verification of textual claims against textual sources. When compared to textual entailment (TE)/natural language inference \citep{Dagan2009,bowman2015large}, the key difference is that  in these tasks the passage to verify each claim %(referred to as premise)
is given, and in recent years it typically consists a single sentence, while in verification systems it is retrieved from a large set of documents in order to form the evidence.
Another related task is question answering (QA), for which approaches have recently been extended to handle large-scale resources such as Wikipedia \citep{Danqi:2017}. 
%AV: removed this for space
%Recent task formulations resembling the evidence retrieval part in verification since they return the textual span answering the question. 
However,  questions typically provide the information needed to identify the answer, while information missing from a claim can often be crucial in retrieving refuting evidence. 
%While recent task formulations define the task as returning the textual span answering the question, thus resembling the evidence retrieval part in verification, the questions typically provide the information needed to identify the answer directly, while information missing from a claim can often be crucial in retrieving refuting evidence. 
For example, a claim stating ``Fiji's largest island is Kauai.'' can be refuted by retrieving ``Kauai is the oldest Hawaiian Island.'' as evidence.%\todo{DO you thing this helps? Could remove it completely if not}
%For example, a claim stating ``Fiji's largest island is Kawelohea.'' can be refuted by using the source sentence ``Fiji's largest island is Viti Levu.''
%AV: KBP (+SP) Let's mention it only as a possible way of making a system
%The extraction of relational facts is a well studied Natural Language Processing discipline: forums and shared tasks such as NIST Text Analysis Conferences (e.g. \newcite{Ji2010}) study extraction of relational facts from documents.

Progress on the aforementioned tasks %, as well as others such as text classification,
has benefited from the availability of large-scale datasets \citep{bowman2015large, rajpurkar2016squad}. However, despite the rising interest in verification and fact checking among researchers, %in natural language processing and machine learning,
the datasets currently used for this task are limited to a few hundred claims. Indicatively, the recently conducted Fake News Challenge \citep{FNC} with 50 participating teams used a dataset consisting of 300 claims verified against 2,595 associated news articles which is orders of magnitude smaller than those used for TE and QA.

In this paper we present a new dataset for claim verification, FEVER: Fact Extraction and VERification. It consists of 185,445 claims manually verified against the introductory sections of Wikipedia pages and classified as \textsc{Supported}, \textsc{Refuted} or \textsc{NotEnoughInfo}. 
 For the first two classes, systems and annotators need to also return the combination of sentences forming the necessary evidence supporting or refuting the claim (see Figure~\ref{fig:ex}). The claims were generated by human annotators extracting claims from Wikipedia and %manually
 mutating them in a variety of ways, some of which were meaning-altering. The verification of each claim was conducted in a separate annotation process by annotators who were aware of the page but not the sentence from which original claim was extracted and thus in 31.75\% of the claims more than one sentence was considered appropriate evidence. %\todo[color=yellow, author=Christos]{Subset of the 16.82 above? JT No. 12.15 of the entire dataset}.
Claims require composition of evidence from multiple sentences in 16.82\% of cases. Furthermore, in 12.15\% of the claims, this evidence was taken from multiple pages.
%Annotators selected multiple sentences as evidence , 
%\todo[color=yellow, author=Christos]{It's not clear that we allowed/ encouraged multiple individual evidence from the same page}
%Furthermore, world knowledge was restricted to Wikipedia pages beyond the one from which the claim was extracted, 

  To ensure annotation consistency, we developed suitable guidelines and user interfaces,
 %to assist the annotators
 resulting in inter-annotator agreement of 0.6841 in
 Fleiss $\kappa$ \citep{fleiss1971measuring} in claim verification classification, and 95.42\% precision and 72.36\% recall in evidence retrieval. 

To characterize the challenges posed by FEVER we develop a pipeline approach which, given a claim, first identifies relevant documents, then selects sentences forming the evidence from the documents and finally classifies the claim w.r.t. evidence. 
The best performing version achieves 31.87\% accuracy in verification when requiring correct evidence to be retrieved for claims \textsc{Supported} or \textsc{Refuted}, and 50.91\% if the correctness of the evidence is ignored, both indicating the difficulty but also the feasibility of the task.
We also conducted oracle experiments in which components of the pipeline were replaced by the gold standard annotations, and observed that the most challenging part of the task is selecting the sentences containing the evidence.
%are sentence selection and correctly differentiating between sentences which are informative from those which are evidential. AV: removed as it is unclear what's informative vs evidential
In addition to publishing the data via our website\footnote{
\url{http://fever.ai}}, we also publish the annotation interfaces\footnote{\url{https://github.com/awslabs/fever}} and the baseline system\footnote{\url{https://github.com/sheffieldnlp/fever-baselines}} to stimulate further research on verification.
%\todo{could make the font in the figure small or footnotesize if needed}

%Our approach to joint fact extraction and verification is unlike existing state of the art works in this domain which each tackle the necessary sub-tasks in a siloed %AV: isolated? might be easier to understand (reviewers are non-native speakers often) manner. 

%Key messages:
%\begin{itemize}
%\item Fact
%\item Sources may not be verified
%\item Not all knowledge can be easily represented in a graph - can text be used instead
%\item Recognizing textual entailment in SLNLI is a very simple task
%\item 
%\end{itemize}

%In today's era of Big Data, the need for semantic information is paramount: from question answering to news aggregation, and from machine translation to bioinformatics, AI systems require vast amounts of structured data. However, s

%automating the construction of Knowledge Bases (KBs) from publicly available documents. 
%KBs are structured stores of facts that enable many natural language technologies such as Question Answering, Semantic Search and Digital Personal Assistants. 

% AV: I would put an example of the task in the first page if possible or top of the second one; always good to the reader to think our way as early as possible.

%Information verification - trending topic. Many instances of fact checking related to fake news

\begin{figure}[t]
    \parskip=0pt
    \begin{framed}
    
    \parskip=0pt
      \begin{description}
\vspace{-0.1in}      
          \item[Claim:] The Rodney King riots took place in the most populous county in the USA. 
\vspace{-0.1in}          
          \item[\texttt{[wiki/Los\_Angeles\_Riots]}] The 1992 Los Angeles riots, \underline{also known as the Rodney King riots} were a series of riots, lootings, arsons, and civil disturbances that \underline{occurred in Los Angeles County}, California in April and May 1992.
\vspace{-0.1in}
\item[\texttt{[wiki/Los\_Angeles\_County]}] Los Angeles County, officially the County of Los Angeles, \underline{is the most populous county in the USA}. 
      \item[Verdict:] Supported

      \end{description}
      
    \parskip=0pt
\vspace{-0.2in}
\end{framed}

  \caption{Manually verified claim %generated by a human,
   requiring evidence from multiple Wikipedia pages.}
  \label{fig:ex}  
\end{figure}

%% file: 2_related_works.tex
%auto-ignore
\section{Related Works}

%when I am talking about SQUAD mention DrQA as an approach that combined DrQA
% connection to SNLI

%AV: Replace this with V&R 2014 which had a dataset, later expanded by Yang into Liar Liar
%AV: TREC task on evidence or something like that?

\citet{Vlachos2014} constructed a dataset for claim verification consisting of 106 claims, selecting data from
fact-checking websites such as PolitiFact, %\footnote{\url{http://www.politifact.com/truth-o-meter/statements/}},
taking advantage of the labelled claims available there.
However, in order to develop claim verification components we typically require the justification for each verdict, including the sources used. While this information is usually available in justifications provided by the journalists, they are not in a machine-readable form. Thus, also considering the small number of claims, the task defined by the dataset proposed remains too challenging for the ML/NLP methods currently available. 
\citet{Wang2017} extended this approach by including all 12.8K claims available by Politifact via its API, however the justification and the evidence contained in it was ignored in the experiments as it was not machine-readable. Instead, the claims were classified considering only the text and the metadata related to the person making the claim. While this rendered the task amenable to current NLP/ML methods, it does not allow for verification against any sources and no evidence needs to be returned to justify the verdicts.

The Fake News challenge \cite{FNC} modelled verification as stance classification: given a claim and an article, predict whether the article supports, refutes, observes (neutrally states the claim) or is irrelevant to the claim. It consists of 50K labelled claim-article pairs, combining 300 claims with 2,582 articles. The claims and the articles were curated and labeled by journalists in the context of the Emergent Project \citep{Silverman2015}, and the dataset was first proposed by \citet{Ferreira2016}, who only classified the claim w.r.t.\ the article headline instead of the whole article. %In both works, even though the claims were labelled by the journalists as True or False, this part of the annotation was not considered since these verdicts were made taking into account information beyond the text, such as the reputation of the articles. 
Similar to recognizing textual entailment (RTE) \cite{Dagan2009}, the systems were provided with the sources to verify against, instead of having to retrieve them. 
%AV: removed for space
%This challenge was also  highlighted for other tasks such as QA and information extraction by \citet{Danqi:2017} and \citet{narasimhan-yala-barzilay:2016} respectively. %AV: Narasimhan used the Gun Violence Database; not sure that qualifies as a corpus per se.

A differently motivated but closely related dataset is the one developed by \citet{angeli2014naturalli}  
to evaluate natural logic inference for common sense reasoning, as it evaluated simple claims such as ``not all birds can fly'' against textual sources --- including Wikipedia --- which were processed with an Open Information Extraction system \citep{Mausam:2012:OLL:2390948.2391009}. However, the claims were small in number (1,378) and limited in variety as they were derived from eight binary ConceptNet relations \citep{TandonDeMeloWeikum2011}.

Claim verification is also related to the multilingual Answer Validation Exercise \citep{Rodrigo2009} conducted in the context of the TREC shared tasks. Apart from the difference in dataset size (1,000 instances per language), the key difference is that the claims being validated were answers returned to questions by QA systems. The questions and the QA systems themselves provide additional context to the claim, while in our task definition the claims are outside any particular context. In the same vein, \citet{kobayashi2017automated} collected a dataset of 412 statements in context from high-school student exams that were validated against Wikipedia and history textbooks.

%% file: 3_new_corpus.tex
\section{Fact extraction and verification dataset}
%In this section we will describe our annotation methodology:

%\subsection{Data Collection}
The dataset was constructed in two stages:% through two separate annotation tasks:
\vspace*{-0.05in}
\begin{description}
\setlength\itemsep{0em}
\item[Claim Generation] Extracting information from Wikipedia %sentences 
and generating claims from it. %as sentences and making mutations to these claims. %AV: Shorter, and avoiding mutations which are not yet explained.
\item[Claim Labeling] Classifying whether %, given a piece of evidence, AV: sounds like we give them the evidence but we don't, they have to find it
a claim is supported or refuted by Wikipedia and selecting the evidence for it, or deciding there’s not enough information to make a decision.
\end{description}
\subsection{Task 1 - Claim Generation}
The objective of this task was to generate %both factual and mutated 
claims %\todo{rephrase it to avoid true? Also not all mutations are meaning altering? Maybe just call them factual claims?} 
from information extracted from Wikipedia. We used the June 2017 Wikipedia dump, processed it with Stanford CoreNLP \cite{manning2014stanford},
%\footnote{
%%To increase the reproducibility of the experiments here,
%and any subsequent ones based on the dataset,
%The original dump and the processed version of %Wikipedia can be downloaded from the website}
 and sampled sentences from the introductory sections of approximately 50,000 popular pages.\footnote{These consisted of 5,000 from a Wikipedia `most accessed pages' list and the pages hyperlinked from them.}
 %\footnote{5000 of these appeared on a Wikipedia `most accessed pages' list. We also included all hyperlinked pages: totaling 50,000.}. %We used  

The annotators were given a sentence from the sample
%from the introductory section of a Wikipedia page 
chosen at random, and were asked to generate a set
of %individual 
\textbf{claims} containing a single piece of information, 
%from the original Wikipedia sentence,
focusing on the entity that its original Wikipedia page was about. We asked the annotators to generate %their
claims about a single fact which could be arbitrarily complex and allowed for a variety of expressions for the entities. %\todo{check: is this on the claims as extracted from wikipedia, or including the mutated ones?... yep: means including the mutated ones??!!. all of them original and mutated, in train, dev, test - JT}
%This can be thought of as extracting a knowledge base \textit{triple} (subject, predicate, object) but in natural language and without resolving the strings to specific KB entities/relations.
%\todo{I see what you mean, but not sure I would call it triple (which are usually simple, or simpler than SNLI's hypotheses). How long are our claims?}
%\todo[color=yellow]{This is how we presented the task to the trainers. Indeed some times the claims end up being longer and more complex than triples, but I would argue that the majority follow this rule.}
%AV: this is about the original sentence
%In the above protocol, 

If only the source sentences were used to generate claims then this would result in trivially verifiable % non-challenging 
claims, % from an information extraction point of view (
as the new claims would in essence be simplifications and paraphrases.
At the other extreme, if we allowed world knowledge to be freely incorporated it would result in claims that would %not
be hard to verify on Wikipedia alone.
We address this issue %gap
by introducing a \textbf{dictionary}: a list of terms that were (hyper-)linked in the original sentence, along with the first sentence from their corresponding Wikipedia pages. Using this dictionary, we provide additional knowledge that can be used to increase the complexity of the generated claims in a controlled manner.

The annotators were also asked to generate \textbf{mutations} of the claims: %distorted %\todo{not all mutations distort the meaning} 
%\todo[color=yellow]{All mutation distort the meaning, even if they leave the truth value unchanged.}
altered versions of the original claims, which may or may not change whether they are supported by Wikipedia, or even if they can be verified against it.
%by making it negative, substituting words or ideas or by making words more or less specific. 
Inspired by the operators used in Natural Logic Inference \citep{angeli2014naturalli}, we specified six types of mutation: paraphrasing, negation, substitution of an entity/relation with a similar/dissimilar one, and making the claim more general/specific. See Appendix~\ref{appdx:guidelines} for the full guidelines. %\todo{provided as supplementary material to this submission?}.

During trials of the annotation task, we discovered that the majority of annotators had difficulty generating non-trivial negation mutations 
(e.g.\ mutations %that did not rely only on 
beyond adding ``not'' to the original). %and claims with dissimilar enough content  
%(e.g.\ for the claim ``Keanu Reeves is an actor'', the mutation ``Keanu Reeves is a director'' is not dissimilar enough since these two roles often co-exist). 
%AV: Not clear to me what dissimilar enough is or should be here given the example. Removed it to save space.
%We also found that annotators would often reuse or mix the different mutations (e.g. generating a negation by substitution or specification). \todo{This mutation mixing sounds non-trivial and desirable?} 
Besides providing numerous examples %providing inspiration 
for each %covering the different 
mutation, we also redesigned the annotation interface so that all mutation types were visible at once and highlighted mutations that contained ``not'' %or ``n't''
in order to discourage trivial negations. Finally, we provided the annotators with an %simplified
ontology diagram %\footnote{Figure 3 in the Appendix.}
to illustrate the different levels of entity similarity and class membership.

This process resulted in claims (both extracted and mutated) with a mean length of 9.4 tokens which is comparable to the average hypothesis length of 8.3 tokens in \newcite{bowman2015large}.

\subsection{Task 2 - Claim Labeling}
\begin{figure*}
\centering
\includegraphics[width=.9\textwidth]{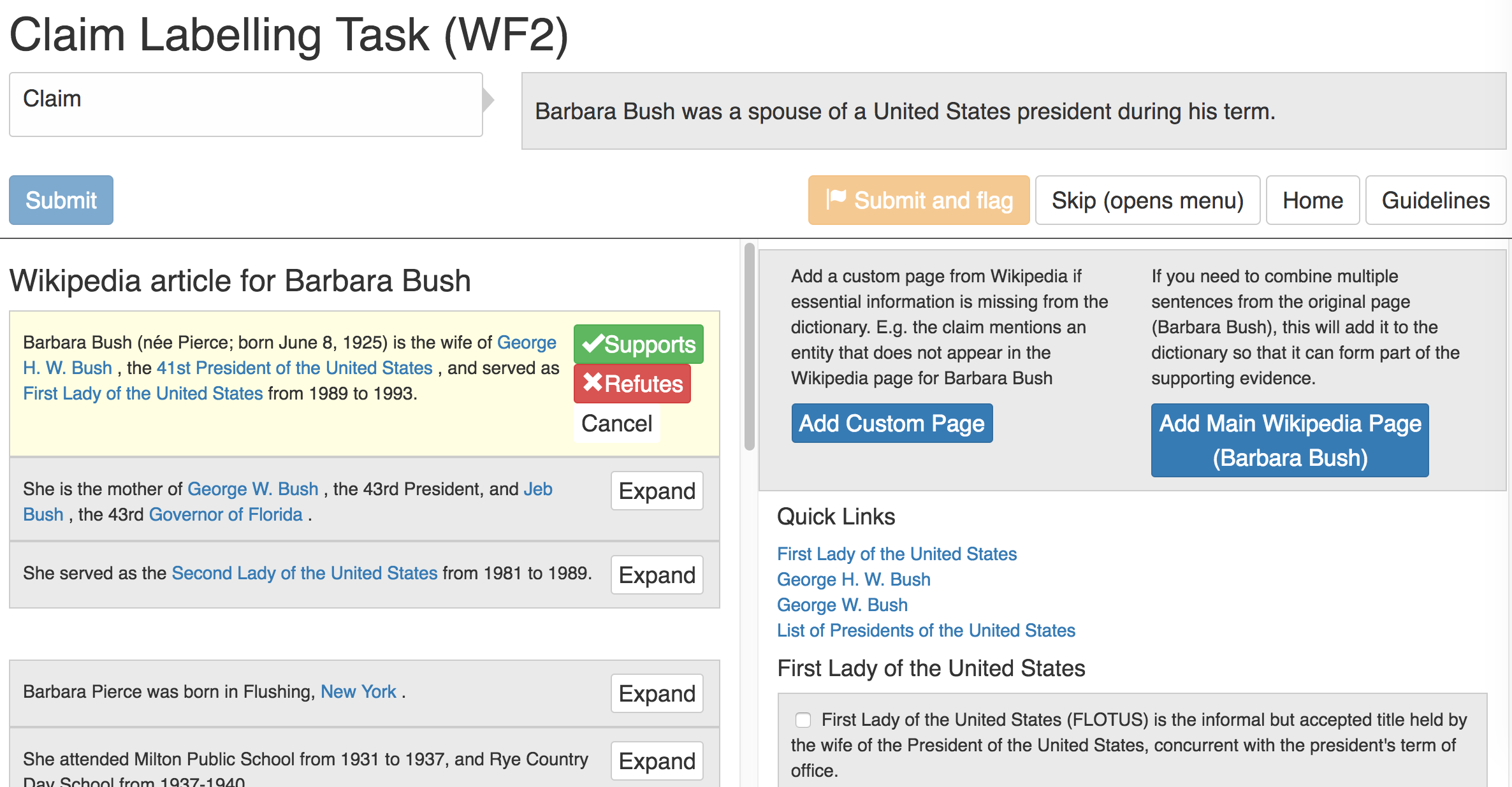}
\caption{Screenshot of Task 2 - Claim Labeling}
\label{fig:task2-screenshot}
\end{figure*}

%\todo{AV: Explain the bit about considering the title of the page as implicitly considered part of the evidence. JT: done}
%The purpose of this task was to identify evidence\todo{Isn't the purpose to classify them using evidence?} from Wikipedia page(s) that can be used to support or refute the individual claims (original or mutated), generated during Task 1. 
The annotators were asked to label each individual claim % (original or mutated),
generated during Task 1 as \textsc{Supported}, \textsc{Refuted} or \textsc{NotEnoughInfo}. For the first two cases, the annotators were asked to find the evidence from any page that supports or refutes the claim (see Figure \ref{fig:task2-screenshot} for a screenshot of the interface). 
In order to encourage inter-annotator consistency, we gave the following general guideline: 
%We wanted to adopt a ``reasonable person'' standard of proof. Here is a quote from the guidelines: %\todo{A bit odd to have it mentioned here, seems quite important, maybe move it? earlier}:
\begin{quotation}
  %As a guide - you should ask yourself:
  If I was given only the selected sentences, do I have strong %\todo[color=yellow]{The word `stronger' is in the guidelines}%\todo{or strong? Say Shakira being Colombian strengthens my belief against her having an Irish passport.} 
  reason to believe the claim is true (supported) or stronger reason to believe the claim is false (refuted). If I'm not certain, what additional information (dictionary) do I have to add to reach this conclusion.
\end{quotation}

In the annotation interface, all sentences from the introductory section of the page for the main entity of the claim and of every linked entity in those sentences 
%\todo[color=yellow]{Rephrased for clarity}
%\todo{hmm, this means the sentence kind of biases the evidence} 
were provided as a default source of evidence (left-hand side in Fig.~\ref{fig:task2-screenshot}). Using this interface the annotators recorded the sentences necessary to justify their classification decisions. 
%that contain relevant supplementary information are then selected %in addition to the original sentence \todo{is the original sentence included by default?} 
 %to record the sum of information used in justifying the annotators' decisions. 
 In order to allow exploration beyond the main and linked pages, we also allowed annotators to add an arbitrary Wikipedia page by providing its URL and the system would add its introductory section as additional sentences that could be then %individually
 selected as evidence (right-hand side in Fig.~\ref{fig:task2-screenshot}). The title of the page could also be used as evidence to resolve co-reference, %or ambiguity,
 but this decision was not explicitly recorded. We did not set a hard time limit for the task, but the annotators were advised not to spend more than 2-3 minutes per claim.
The label \textsc{NotEnoughInfo} was used if the claim could not be supported or refuted by any amount of information in Wikipedia (either because it is too general, or too specific). 

\subsection{Annotators}
The annotation team consisted of a total of 50 members, 25 of which were involved only in the first task. % (see below).
All annotators were native US English speakers and were trained either directly by the authors, or by experienced annotators. The interface for both tasks was developed by the authors in collaboration with an initial team of two annotators. Their notes and suggestions were incorporated into the annotation guidelines.
%Both the interface and the annotation guidelines went through a number of iterations to increase usability, as well as inter-annotator agreement on an initial set of claims that were later discarded.\todo{sound odd, maybe remove?} 

The majority of the feedback received from the annotators was very positive: they found the task engaging and challenging, and after the initial stages of annotation they had developed an understanding of the needs of the task which let them discuss %and come up with
solutions about edge cases as a group. %, increasing their inter-annotator agreement. %AV: I wouldn't say this without backing it with evidence (would be great if we could!)

\subsection{Data Validation}
Given the complexity of the second task (claim labeling), we conducted three forms of data validation: % . We evaluated the validity of the data using three separate methods: 
5-way inter-annotator agreement, agreement against \textit{super-annotators} (defined in Section~\ref{sec:eval_annotators})%\todo[color=yellow, author=Christos]{Add definition here.}\todo[color=green, author=James]{Added forward pointer to section 3.4.2}
, and manual validation by the authors. 
The validation for claim generation was done implicitly during claim labeling. As a result 1.01\% of claims were skipped, 2.11\% contained typos and 6.63\% of the generated claims were flagged 
as too vague/ambiguous and were excluded e.g.\  %``Chris Evans worked.''.
%or ``The Godfather Part II is a  work.'', 
or ``Sons of Anarchy premiered.''.
% \todo[color=yellow, author=Christos]{Mention the results of the second round of annotation we did (currently in section 5.8)}
%\todo{How do you mean by the unverifiable label? We should mention it Task2}

\subsubsection{5-way Agreement}
We randomly selected 4\% ($n=7506$) of claims which were not skipped to be annotated by 5 annotators. We calculated the Fleiss $\kappa$ score \cite{fleiss1971measuring} to be 0.6841 which we consider encouraging %is impressive 
given the complexity of the task. In comparison \citet{bowman2015large} reported a $\kappa$ of $0.7$ for a simpler task, since the annotators were  given the premise/evidence to verify a hypothesis against without the additional task of finding it.%\todo{Added this to justify the impressive}

\subsubsection{Agreement against \textit{Super-Annotators}} %AV: Keep quotes for strings from the examples
\label{sec:eval_annotators}
We randomly selected 1\% of the data to be annotated by \textit{super-annotators}: expert annotators with no suggested time restrictions. 
The purpose of this exercise was to provide as much coverage of evidence as possible. We instructed the \textit{super-annotators} to search over the whole Wikipedia for %any and all
every possible sentence that could be used as evidence. We compared the regular annotations against this set of evidence and the precision/recall was 95.42\% and 72.36\% respectively. %\todo{Why average? @Christo, my precision/recall script did global P/R and a daily/weekly breakdown. I don't think it computes average - unless its changed. JT}
%against the super-annotators' evidence selection  \todo{thus validating our annotation interface choices?}

% \begin{figure}
% \centering
% \includegraphics[width=\columnwidth]{annotation-figures/prec_by_user}
% \caption{Task 2 precision by user (scored against the oracles) in ascending order. All except two annotators achieve $> 90$\% precision.}
% \label{fig:task2-prec-by-user}
% \end{figure}

\subsubsection{Validation by the Authors}
As a final quality control step, we chose 227 examples and annotated them for accuracy of the labels and the evidence provided. We found that 91.2\% of the examples were annotated correctly. 3\% of the claims were mistakes in claim generation that had not been flagged during labeling. We found a similar number of these claims which did not meet the guidelines during a manual error analysis of the baseline system (Section~\ref{sec:error_analysis}).

%  , and during the re-annotation by the authors, 3\% were marked as such.
%\todo{both for label and evidence? To be honest when there was a lot of evidence I didn't check them all.}

\subsubsection{Findings}
%The main finding concerning the first task is that some of the claims were too vague (mostly due to the generalization mutation). For instance, claims like ``Chris Evans worked.'', or ``The Godfather Part II is a work.'', or ``Sons of Anarchy premiered.'' 7.6\% were flagged by the annotators of the second task as vague/ambiguous (and were excluded)\todo{maybe mention it in WF2?}, and during the re-annotation by the authors, 3\% were marked as such.

When compared against the \textit{super-annotators}, all except two annotators achieved $> 90$\% precision and all but 9 achieved recall $> 70$\% in evidence retrieval. The majority of the low-recall cases are for claims such as ``Akshay Kumar is an actor.'' where the \textit{super-annotator} added 34 sentences as evidence, most of them being filmography listings (e.g. ``In 2000, he starred in the Priyadarshan-directed comedy Hera Pheri'').

During the validation by the authors, we found that most of the examples that were annotated incorrectly were cases where the label was correct, but the evidence selected was not sufficient (only 4 out of 227 examples were labeled incorrectly according to the guidelines). 
%For example, a claim ``[Person] was in [film]'' can be supported with a sentence stating that ``[Person] is starring in [film]''. However, for the claim ``The Eagles are a band from Indiana'', a sentence that says that they are from California could only be used as evidence of refutation if a sentence that mentions Indiana and California are different states is also selected.\todo[color=purple, author=Andreas]{Have we adhered to this? My sense is that we relaxed this a bit. Also in the paragraph below.} 

%We noticed the above mentioned disagreement since our initial iterations with the annotators. While we could not define an absolute standard, 
We tried to resolve % mitigate
this issue %disagreement. We asked 
by asking our annotators to err on the side of caution. For example, while the claim ``Shakira is Canadian'' could be labeled as \textsc{Refuted} by the sentence ``Shakira is a Colombian singer, songwriter, dancer, and record producer'', we advocated that unless more explicit evidence is provided (e.g. ``She was denied Canadian citizenship''), the claim should be labeled as \textsc{NotEnoughInfo}, since dual citizenships are permitted, and the annotators' world knowledge should not be factored in. 
%This contributed to the large number of claims (26.2\%) annotated as \textsc{NEI}.\todo{I would remove this statement; sounds like as something that shouldn't have happened, which is not what we mean here.}

A related issue is entity resolution. For a claim like ``David Beckham was with United.'', it might be trivial for an annotator to accept ``David Beckham made his European League debut playing for Manchester United.'' as supporting evidence. This implicitly assumes that ``United'' refers to ``Manchester United'', however there are many Uniteds in Wikipedia and not just football clubs, e.g.\ United Airlines. %for instance a system can use the roster of United Airlines pilots as evidence against the claim).
The annotators knew the page of the main entity % and the pages within the immediate neighborhood, 
and thus it was relatively easy to resolve ambiguous entities.
While we provide this information as part of the dataset, we argue that it should only be used for system training/development.%\todo{Given space we can add that if during testing a system returns a correctly supported label differing from the one in our dataset, we would consider it correct during the manual evaluation.}
%, we argue that during testing it should be ignored,.

%% file: 4_baseline.tex
%auto-ignore
\section{Baseline System Description}
\label{sec:baselines}
%\todo[inline]{The task requires participants to extract evidence from wikipedia from }

We construct a simple pipelined system comprising three components: document retrieval, sentence-level evidence selection and textual entailment. Each component is evaluated in isolation through oracle evaluations on the development set and we report the final %pipeline scores.
accuracies on the test set. %To highlight the important of the sentence-level evidence selection in our task, we complete an evaluation with the sentence selection module removed. %AV: space saving

%Describe how a symbolic approach could be done too (maybe not in this section)

\paragraph{Document Retrieval}
We use the document retrieval component from the DrQA system \cite{Danqi:2017} which returns the $k$ nearest documents for a query using cosine similarity between binned unigram and bigram Term Frequency-Inverse Document Frequency (TF-IDF) vectors.

\paragraph{Sentence Selection}
Our simple sentence selection method ranks sentences by TF-IDF similarity to the claim.
We sort the most-similar sentences first and tune a cut-off using validation accuracy on the development set.  We evaluate both DrQA and a simple unigram TF-IDF implementation to rank the sentences for selection. We further evaluate impact of sentence selection on the RTE module by predicting entailment given the original documents without sentence selection. 

\paragraph{Recognizing Textual Entailment}
\label{sec:strat}
We compare two models for recognizing textual entailment. For a simple well-performing
baseline, we selected \citet{riedel2017fnc}'s submission from the 2017 Fake News Challenge. %\cite{FNC}. AV: space saving
It is a multi-layer perceptron (\textbf{MLP}) with a single hidden layer which uses term frequencies and TF-IDF cosine similarity between the claim and evidence as features. 

Evaluating the state-of-the-art in RTE, we used a decomposable attention (\textbf{DA}) model between the claim and the evidence passage \cite{Parikh2016}. We selected it because at the time of development this model was the highest scoring system for the Stanford Natural Language Inference task \cite{bowman2015large} with publicly available code that did not require the input text to be %constituency
parsed syntactically, nor was an ensemble.

The RTE component must correctly classify a claim as {\sc NotEnoughInfo} when the evidence retrieved %a page 
is not relevant or informative. However, the instances labeled as \textsc{NotEnoughInfo} have no evidence annotated, thus cannot be used to train RTE for this class. To overcome this issue, we simulate training instances for the {\sc NotEnoughInfo} through two methods: sampling a sentence from the nearest page %\todo{(RESOLVED-JT?)you use the same for both the sentence and the doc-level training, right? Call it NearestN (nearest neighbor). Not quite... I sample a sentence from the nearest page, or from a random page}
(\textsc{NearestP}) to the claim as evidence using our document retrieval component and sampling a sentence from Wikipedia uniformly at random (\textsc{RandomS}).

%% file: 5_evaluation.tex
%auto-ignore
\section{Experiments}

\subsection{Dataset Statistics}
%AV: Made this a subsection of the annotation
We partitioned the annotated claims into training, development and test sets. We ensured that each Wikipedia page used to generate claims occurs in exactly one set. 
We reserved a further 19,998 examples for use as a test set for a shared task.

% \begin{table}[h]
% \centering
% \begin{tabular}{|c|c|c|c|}\hline
% \textbf{Split} & \textsc{Supported} & \textsc{Refuted} & \textsc{NEI} \\ \hline 
% Training & 57,912 & 22,584 & 24,897 \\
% Dev & 3,333 & 3,362 & 3,304 \\
% Test & 3,333 & 3,363 & 3,303 \\
% \hline
% \end{tabular}
\begin{table}[h]
\centering
\begin{tabular}{@{}rccc@{}}
\toprule
\textbf{Split} & \textsc{Supported} & \textsc{Refuted} & \textsc{NEI} \\ \hline 
Training & 80,035 & 29,775 & 35,639 \\
Dev & 3,333 & 3,333 & 3,333 \\
Test & 3,333 & 3,333 & 3,333 \\
Reserved & 6,666 & 6,666 & 6,666 \\
\bottomrule
\end{tabular}

\caption{Dataset split sizes for \textsc{Supported}, \textsc{Refuted} and \textsc{NotEnoughInfo} (\textsc{NEI}) classes}

\end{table}
\subsection{Evaluation}

Predicting whether a claim is \textsc{Supported}, \textsc{Refuted} or \textsc{NotEnoughInfo} is a 3-way classification task that we evaluate using accuracy.
In the case of the first two classes, appropriate evidence must be provided, at a sentence-level, to justify the classification. 
 We consider an answer returned correct for the first two classes only if correct evidence is returned. %, in the form of sentence(s). 
Given that the development and test datasets have %been designed with 
balanced class distributions, a random baseline will have $\sim$ 33\% accuracy
%score $\frac{1}{3}$, %AV: avoid "score", tends to be ambiguous 
if one ignores the requirement for evidence for \textsc{Supported} and \textsc{Refuted}.

We evaluate the correctness of the evidence retrieved by computing the $F_1$-score of all the predicted sentences in comparison to the human-annotated sentences for those claims requiring evidence on our complete pipeline system (Section~\ref{sec:full_pipeline}). As in Fig.~\ref{fig:ex}, 
some claims require multi-hop inference %over evidence 
involving sentences from more than one document to be correctly supported as \textsc{Supported}/\textsc{Refuted}. In this case all sentences must be selected for the evidence to be marked as correct. We report this as the proportion of \textit{fully supported claims}.
Some claims may be equally supported by different pieces of evidence; in this case one complete set of sentences should be predicted. 

Systems that select information that the annotators did not will be penalized in terms of precision.
%through the precision portion of the $F_1$-score.
We recognize that it is not feasible to ensure that the evidence selection annotations are complete, %\todo[color=yellow]{Add a pointer to say that it's pretty good given 74.84 recall}
 nevertheless we argue that they are useful for automatic evaluation during system development. For a more reliable evaluation we advocate %Incompleteness will be mitigated by
crowd-sourcing annotations of false-positive predictions at a later date in a similar manner to the TAC KBP Slot Filler Validation \cite{Ellis2016}. %\todo{Some description of oracle evaluation will be better here}

%\paragraph{Oracle design}\todo{Not sure I get this paragraph, maybe just have each oracle explain in its respective section?} In completing oracle evaluations of the baseline document retrieval and sentence selection components, we additionally evaluate recall.  This is under the working assumption that the textual entailment classifier will accurately predict \textsc{NotEnoughInfo} for evidence that cannot be used to support or refute a claim. In addition to the per-claim recall metric, we show per-document/per-sentence for \textsc{Supported} / \textsc{Refuted} claims to highlight the effects our requiring the minimal set of evidence in comparison to requiring all evidence. In evaluation of the RTE component, we report accuracy without the requirement for correct evidence.

\subsection{Document Retrieval}
\label{sec:dr}
The document retrieval component of the baseline system returns the $k$ nearest documents to the claim using the DrQA \cite{Danqi:2017} TF-IDF implementation to return the $k$-nearest documents.
In the scenario where evidence from multiple documents is required, $k$ must be greater than this figure.
We simulate the upper bound in accuracy using an oracle 3-way RTE classifier that  predicts \textsc{Supported}/\textsc{Refuted} ones correctly only if the documents containing the supporting/refuting evidence are returned by document retrieval and
always predicts \textsc{NotEnoughInfo} instances correctly independently of the evidence. Results are shown in Table~\ref{tab:retrieval}.
%given the retrieved documents by re-weighting the recall and assigning a perfect score for the \textsc{NotEnoughInfo} class which makes up the remaining third of the development and test datasets: $Oracle(C) = \frac{1}{3}+\frac{2}{3}Recall(C)$. \todo{Why we call this as accuracy in Table 2}

\begin{table}[h]

\centering
% \begin{tabular}{|c||c|c||c|}\hline

% \multirow{2}{*}{$\mathbf{k}$} & \multicolumn{2}{c||}{\textbf{Recall}} & \textbf{Accuracy} \\
% & \textbf{Per-Claim} & \textbf{Per-Doc} & \textbf{Oracle} \\ \hline 
% 1 & 0.3072 & 0.2811 & 0.5381 \\
% 5 & 0.5883 & 0.5530 & 0.6255 \\
% 10 & 0.6811 & 0.6447 & 0.7874 \\
% 25 & 0.7778 & 0.7426 & 0.8519\\
% 100 & 0.8763 & 0.8461 & 0.9175 \\
% \hline
% \end{tabular}
\begin{tabular}{@{}rccc@{}}
\toprule
\multirow{2}{*}{$\mathbf{k}$} & \multicolumn{1}{c}{\textbf{Fully}} & \textbf{Oracle} \\
& \textbf{Supported (\%)} & \textbf{Accuracy (\%)} \\ 
\midrule 
1 & 25.31 & 50.21 \\
5 & 55.30 & 70.20 \\
10 & 65.86 & 77.24 \\
25 & 75.92 & 83.95 \\
50 & 82.49 & 90.13\\
100 & 86.59 & 91.06 \\
\bottomrule
\end{tabular}
\caption{Dev. set document retrieval evaluation.}
\label{tab:retrieval}
\end{table}

%\todo[inline]{Per-claim recall reduces the task difficulty. But the task is still difficult than SQuAD. For example - DrQA was 78.0\% for SQuAD k=5 on squad.}

\subsection{Sentence Selection}
\label{sec:ss}
Mirroring document retrieval, 
we extract the top $l$-most similar sentences from the $k$-most relevant documents using TF-IDF vector similarity. We modified document retrieval component of DrQA \citep{Danqi:2017} to select sentences using bigram TF-IDF with binning and compared this to a simple unigram TF-IDF implementation using NLTK \citep{Loper2002}. Using the parameters $k=5$ documents and $l=5$ sentences, $55.30\%$ of claims (excluding \textsc{NotEnoughInfo}) can be fully supported or refuted by the retrieved documents before sentence selection (see Table~\ref{tab:retrieval}). After applying the sentence selection component, $44.22\%$ of claims can be fully supported using the extracted sentences with DrQA and only $34.03\%$ with NLTK. This would yield oracle accuracies of $62.81\%$ and $56.02\%$ respectively.

\subsection{Recognizing Textual Entailment}
\label{sec:rte}
The RTE component is trained on labeled claims paired with sentence-level. Where multiple sentences are required as evidence, the strings are concatenated. As discussed in Section~\ref{sec:baselines}, such data is not annotated for claims labeled \textsc{NotEnoughInfo}, thus we compare random sampling-based and similarity-based strategies for generating it. 
%training data for claims labeled \textsc{NotEnoughInfo} as introduced in Section~\ref{sec:strat}.
We evaluate classification accuracy on the development set in an oracle evaluation, assuming correct evidence sentences are selected (Table~\ref{table:oracle}).  
Additionally, for the DA model, we predict entailment given evidence, using the AllenNLP \cite{Gardner2017} pre-trained Stanford Natural Language Inference (SNLI) model for comparison.

%\todo{Say that we used the pretrained allennlp SNLI model and ran it using out dataset}

\begin{table}[h]

\centering
% \begin{tabular}{|c|c|c|c|}\hline
% \multirow{2}{*}{\textbf{Model}} & \multicolumn{3}{c|}{\textbf{Accuracy}}  \\
% & \textsc{NearestP} & \textsc{RandomS} & \textsc{SNLI} \\ \hline
% MLP &  0.6285 & 0.7188 & - \\
% LSTM & 0.7797  & 0.8639 & 0.4080 \\
% \hline
% \end{tabular}
\begin{tabular}{@{}cccc@{}}
\toprule
\multirow{2}{*}{\textbf{Model}} & \multicolumn{3}{c}{\textbf{Accuracy (\%)}}  \\
\cmidrule{2-4}
& \textsc{NearestP} & \textsc{RandomS} & \textsc{SNLI} \\ 
\midrule
MLP &  65.13 & 73.81 & - \\
DA & 80.82 & 88.00 & 38.54 \\
\bottomrule
\end{tabular}

\caption{Oracle classification on claims in the development set using gold sentences as evidence}
\label{table:oracle}
\end{table}

The random sampling (\textsc{RandomS}) approach (where a sentence is sampled at random from Wikipedia in place of evidence for claims labeled as \textsc{NotEnoughInfo}) yielded sentences that were not only semantically different to the claim, but also unrelated. While the the accuracy of models trained with sampling approach is higher in oracle evaluation setting, this may not yield a better system in the pipeline setting. In contrast, the nearest page (\textsc{NearestP}) method samples a sentence from the highest-ranked page returned by our document retrieval module. This simulates finding related information that may not be sufficient to support or refute a claim. We will evaluate both \textsc{RandomS} and \textsc{NearestP} in the full pipeline setting, but we will not pursue the SNLI-trained model further as it performed substantially worse.

\subsection{Full Pipeline}
\label{sec:pipeline}
The complete pipeline consists of the DrQA document retrieval module (Section~\ref{sec:dr}), DrQA-based sentence retrieval module (Section~\ref{sec:ss}), and the decomposable attention RTE model (Section~\ref{sec:rte}). The two parameters: $k$, describing the number documents and $l$, describing the number sentences to return were found using grid-search optimizing the RTE accuracy with the DA model. For the pipeline, we set $k=5$ and $l=5$ and report the development set accuracy, both with and without the requirement to provide correct evidence for the \textsc{Supported}/\textsc{Refuted} predictions (marked as $\mathbf{ScoreEv}$ and $\mathbf{NoScoreEv}$ respectively).

\begin{table}[h]
\centering
% \begin{tabular}{|c|c|c|}\hline
% \multirow{2}{*}{\textbf{Model}} & \multicolumn{2}{c|}{\textbf{Accuracy}} \\
% & $\mathbf{NoScoreEv}$ & $\mathbf{ScoreEv}$ \\ \hline
% MLP / \textsc{NP} & 0.4113 & 0.1531  \\
% MLP / \textsc{RS} & 0.3996 & 0.1577  \\
% LSTM / \textsc{NP} & 0.4845 & \textbf{0.2672} \\
% LSTM / \textsc{RS} & 0.4922 & 0.1954 \\
% \hline
% \end{tabular}

\begin{tabular}{@{}ccc@{}}
\toprule
\multirow{2}{*}{\textbf{Model}} & \multicolumn{2}{c}{\textbf{Accuracy (\%)}} \\
\cmidrule{2-3}
& $\mathbf{NoScoreEv}$ & $\mathbf{ScoreEv}$ \\ 
\midrule
MLP / \textsc{NP} & 41.86 & 19.04  \\
MLP / \textsc{RS} & 40.63 & 19.42  \\
DA / \textsc{NP} & 52.09 & \textbf{32.57} \\
DA / \textsc{RS} & 50.37 & 23.53 \\
\bottomrule
\end{tabular}

\caption{Full pipeline results on development set}
\label{tab:full}
\end{table}

The decomposable attention model trained with \textsc{NearestP} is the most accurate when evidence is considered. Inspection of the confusion matrices shows that the \textsc{RandomS} strategy harms recall for the \textsc{NotEnoughInfo} class. This is due to the difference %in sentence similarities in 
between the sampled pages in the training set and the ones retrieved in the development set causing related but uninformative evidence to be misclassified as \textsc{Supported} and \textsc{Refuted}.
%\todo[color=purple, author=Andreas]{Idea for the future: combine them? Both seem to have merits} 

\paragraph{Ablation of the sentence selection module} We evaluate the impact of the sentence selection module on both the RTE accuracy by removing it.  While the sentence selection module may improve accuracy in the RTE component, it is discarding sentences that are required as evidence to support claims, harming performance (see Section~\ref{sec:ss}). We assess the accuracies in both oracle setting (similar to Section~\ref{sec:rte}) (see Table~\ref{table:oracle_noss}) as well as in the full pipeline (see Table~\ref{table:full_noss}).

In the oracle setting, the decomposable attention models are worst affected by removal of the sentence selection module: exhibiting an substantial decrease in accuracy.  The \textsc{NearestP} training regime exhibits a $17\%$ decrease and the \textsc{RandomS} accuracy decreases by $19\%$, despite near-perfect recall of the \textsc{NotEnoughInfo} class.  

\begin{table}[h]
\centering
% \begin{tabular}{|c|c|c|}\hline
% \multirow{2}{*}{\textbf{Model}} & \multicolumn{2}{c|}{\textbf{Accuracy}}  \\
% & \textsc{NearestP} & \textsc{RandomS} \\ \hline
% MLP &  0.5417 & 0.7204 \\
% LSTM & 0.6066  & 0.8246 \\
% \hline
% \end{tabular}
\begin{tabular}{@{}ccc@{}}
\toprule
\multirow{2}{*}{\textbf{Model}} & \multicolumn{2}{c}{\textbf{Oracle Accuracy (\%)}}  \\
\cmidrule{2-3}
& \textsc{NearestP} & \textsc{RandomS} \\ 
\midrule
MLP &  57.16 & 73.36 \\
DA & 63.68  & 69.05 \\
\bottomrule
\end{tabular}

\caption{Oracle accuracy on claims in the dev. set using gold documents as evidence  (c.f. Table~\ref{table:oracle}).}
\label{table:oracle_noss}
\end{table}

In the pipeline setting, we run the RTE component without sentence selection using $k=5$ most similar predicted documents. The removal of the sentence selection component decreased the accuracy (\textsc{NoScoreEv}) approximately $10\%$ for both decomposable attention models.

\begin{table}[h]
\centering
% \begin{tabular}{|c|c|c|}\hline
% \multirow{2}{*}{\textbf{Model}} & \multicolumn{2}{c|}{\textbf{Accuracy}}  \\
% & \textsc{NearestP} & \textsc{RandomS} \\ \hline
% MLP & 0.4036 & 0.4050  \\
% LSTM & 0.4367 & 0.4334 \\
% \hline
% \end{tabular}
\begin{tabular}{@{}ccc@{}}
\toprule
\multirow{2}{*}{\textbf{Model}} & \multicolumn{2}{c}{\textbf{Accuracy (\%)}}  \\
\cmidrule{2-3}
& \textsc{NearestP} & \textsc{RandomS} \\ 
\midrule
MLP & 38.85 & 40.45  \\
DA & 41.57 & 40.62 \\
\bottomrule
\end{tabular}

\caption{Pipeline accuracy on  the dev. set without the sentence selection module (c.f. Table~\ref{tab:full}).}

\label{table:full_noss}
\end{table}

\subsection{Evaluating Full Pipeline on Test Set}
\label{sec:full_pipeline}
We evaluate our pipeline approach on the test set based on the results observed in Section~\ref{sec:pipeline}. First, we use DrQA to select select 5 documents nearest to the claim. Then, we select 5 sentences using our DrQA-based sentence retrieval component and concatenate them. Finally, we predict entailment using the Decomposable Attention model trained with the \textsc{NearestP} strategy. The classification accuracy is $31.87\%$. Ignoring the requirement for correct evidence ($\mathbf{NoScoreEv}$) the accuracy is $50.91\%$, which highlights that while the systems were predicting the correct label, the evidence selected was different to that which the human annotators chose. The recall of the document and sentence retrieval modules for claims which required evidence on the test set was $45.89\%$ (considering complete groups of evidence) and the precision $10.79\%$. The resulting $F_1$ score is $17.47\%$.

\subsection{Manual Error Analysis}
\label{sec:error_analysis}
Using the predictions on the test set, we sampled $961$ of the predictions with an incorrect label or incorrect evidence and performed a manual analysis (procedure described in Appendix~\ref{appdx:manual}). Of these, $28.51\%$ ($n=274$) had the correct predicted label but did not satisfy the requirements for evidence. The information retrieval component of the pipeline failed to identify any correct evidence in $58.27\%$ ($n=560$) of cases which accounted for the large disparity between accuracy of the system when evidence was and was not considered. Where suitable evidence was found, the RTE component incorrectly classified $13.84\%$ ($n=133$) of claims.

The pipeline retrieved new evidence that had not been identified by annotators in $21.85\%$ ($n=210$) of claims. This was in-line with our expectation given the measured recall rate of annotators (see Section~\ref{sec:eval_annotators}), who achieved recall of $72.36\%$ of evidence identified by the super-annotators.

We found that $4.05\%$ ($n=41$) of claims did not %fully
meet our guidelines. Of these, there were $11$ claims which could be checked without evidence as these either tautologous or self-contradictory. Some correct claims appeared ungrammatical due to the mis-parsing of named entities (e.g. \emph{Exotic Birds} is the name of a band but could be parsed as a type of animal). Annotator errors (where the wrong label was applied) were present in $1.35\%$ ($n=13$) of incorrectly classified claims. 

Interestingly, our system found new evidence that contradicted the gold evidence in $0.52\%$ ($n=5$) of cases. This was caused either by entity resolution errors or by inconsistent information present in Wikipedia pages (e.g. Pakistan was described as having both the $41$st and $42$nd largest GDP in two different pages).

\subsection{Ablation of Training Data}
\begin{figure}[h]
\includegraphics[width=\linewidth]{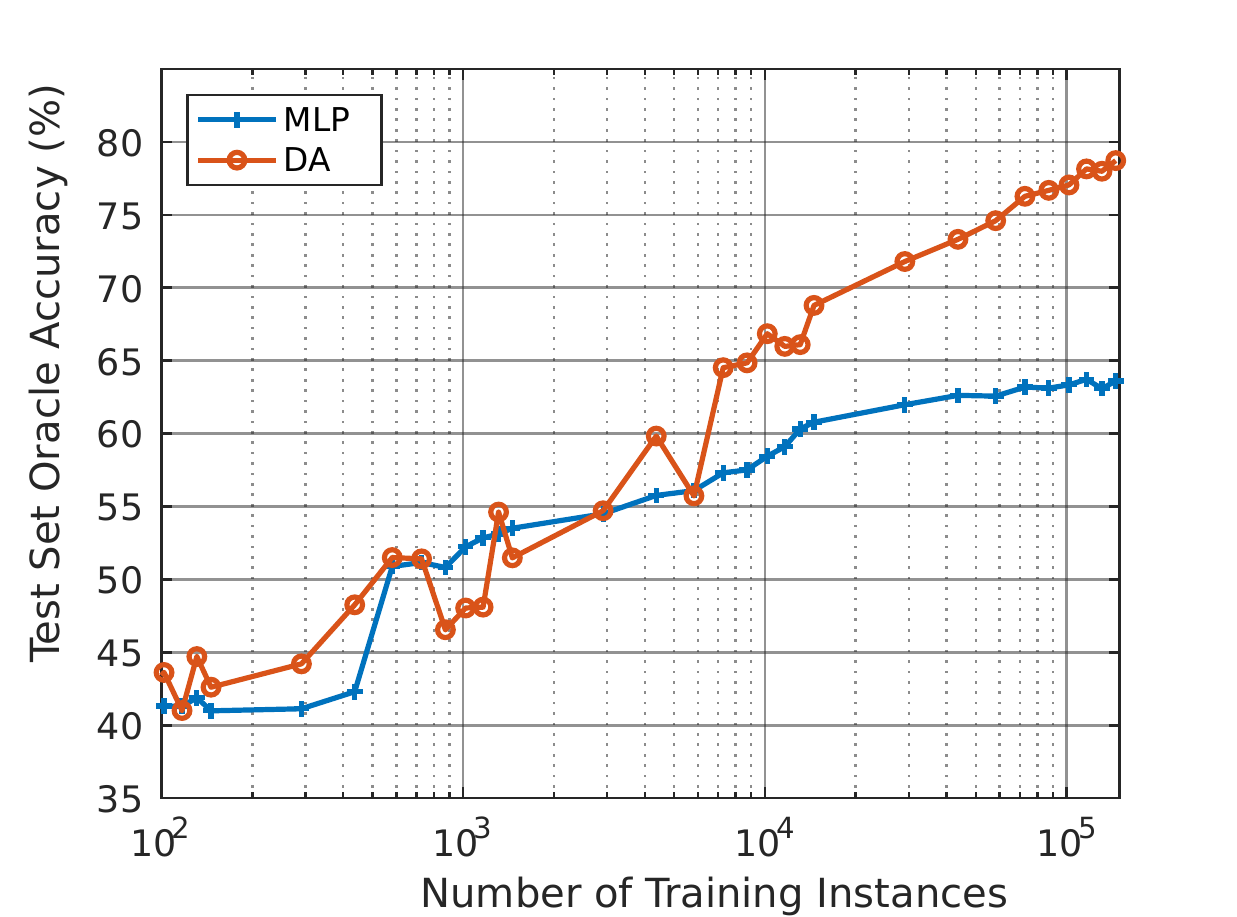}
\caption{Learning curves for the RTE models.}
\label{fig:learning}
\end{figure}

To evaluate whether the size of the dataset is suitable for training the RTE component of the pipeline, we plot the learning curves for the DA and MLP models (Fig.~\ref{fig:learning}). For each model, we trained 5 models with different random initializations using the \textsc{NearestP} method (see Section~\ref{sec:rte}). We selected the highest performing model when evaluated on development set and report the oracle RTE accuracy on the test set. We observe that with fewer than 6000 training instances, the accuracy of DA is unstable. However, with more data, its accuracy increases with respect to the $\log$ of the number of training instances and exceeds that of MLP. While both learning curves exhibit the typical diminishing return trends, they indicate that the dataset is large enough to demonstrate the differences of models with different learning capabilities.

%% file: 6_discussion.tex
%auto-ignore
\section{Discussion}

The pipeline presented and evaluated in the previous section is one possible approach to the task proposed in our dataset, but we envisage different ones to be equally valid and possibly better performing. For instance, 
it would be interesting to test how approaches similar to natural logic inference \cite{angeli2014naturalli} can be applied, where a knowledge base/graph is constructed by reading the textual sources and then a reasoning process over the claim is applied, possibly using recent advances in neural theorem proving \cite{rocktaschel2017end}. A different approach could be to consider a combination of question generation \citep{Heilman:2010:GQS:1857999.1858085} followed by a question answering model such as BiDAF \citep{DBLP:journals/corr/SeoKFH16}, possibly requiring modification as they are designed to select a single span of text from a document rather than return one or more sentences as per our scoring criteria. The sentence-level evidence annotation in our dataset will help develop models selecting and attending to the relevant information from multiple documents and non-contiguous passages. Not only will this enhance the interpretability of predictions, but also facilitate the development of new methods for reading comprehension. %\cite{Kocisky2017}. AV: Feels odd to use it as a reference for reading comprehension, let's not talk about it.
%\todo{JT andreas, I've added the NarrativeQA citation here and discussed getting it right for the wrong reason earlier in the evaluation section}

%AV say: Our dataset is constructed to jointly address open-domain Information Extraction and Fact Checking using either a symbolic or distributional approach to reasoning
%AV: actually, let's leave this, as it would get us in a discussion about what is distributional and what is not, and all the thing in-between.

Another use case for the FEVER dataset is claim extraction: generating short concise textual facts from longer encyclopedic texts. For sources like Wikipedia or news articles, the sentences can contain multiple individual claims, making them not only difficult to parse, but also hard to evaluate against evidence. During the construction on the FEVER dataset, we allowed for an extension of the task where simple claims can be extracted from multiple complex sentences.
%While we haven't examined this aspect of the dataset in this paper, we believe that the extracted claims paired with their original sentences will be a very useful source of data for this task. %AV: Space saving

%Focused summarization

%Links to interpretability: one could train an attention model %JT backs up why we did sentence-level  selection 

Finally, we would like to note that while we chose Wikipedia as our textual source, we do not consider it to be the only source of information worth considering in verification, hence not using {\sc True} or {\sc False} in our classification scheme. We expect systems developed on the dataset presented to be portable to different textual sources.%, and our annotation guidelines and insights to be useful to other verification annotation efforts.

%In future versions of this dataset, we would like to add more sources like news articles and social media posts, to include cross-source verification, and to expand on languages other than English. AV: Space saving

%% file: 7_conclusions.tex
%auto-ignore
\section{Conclusions}
In this paper we have introduced FEVER, a publicly available dataset for fact extraction and verification against textual sources. We discussed the data collection and annotation methods and shared some of the insights obtained during the annotation process that we hope will be useful to other large-scale annotation efforts. 

In order to evaluate the challenge this dataset presents, we developed a pipeline approach that comprises information retrieval and textual entailment components. We showed that the task is challenging yet feasible, with the best performing system achieving an accuracy of 31.87\%.

We also discussed other uses for the FEVER dataset and presented some further extensions that we would like to work on in the future. We believe that FEVER will provide a stimulating challenge for claim extraction and verification systems.

%% file: a_appendix.tex
%auto-ignore
\section{Annotation Guidelines}
\label{appdx:guidelines}
\subsection{Task 1 Definitions}
\paragraph{Claim} A claim is a single sentence expressing information (true or mutated) about a single aspect of one target entity. Using only the source sentence to generate claims will result in simple claims that are not challenging. But, allowing world knowledge to be incorporated is too unconstrained and will result in claims that cannot be evidenced by this dataset. We address this gap by introducing a dictionary that provides additional knowledge that can be used to increase the complexity of claims in a controlled manner.
\paragraph{Dictionary} Additional world knowledge is given to the annotator in a dictionary. This allows for more complex claims to be generated in a structured manner by the annotator. This knowledge may be incorporated into claims or may be needed when labelling whether evidence supports or refutes claims.
\paragraph{Mutation} True claims will be distorted or mutated as part of the claim generation workflow. This may be achieved by making the sentence negative, substituting words or ideas or by making words more or less specific. The annotation system will select which type of mutation will be used.

\noindent\textbf{Requirements:}
\begin{itemize}
\item Claims must reference the target entity directly and avoid use of pronouns/nominals (e.g. he, she, it, the country)
\item Claims must not use speculative/cautious/vague language (e.g. may be, might be, it is reported that)
\item True claims should only be facts that can be deduced by information given in the source sentence and dictionary
\item Minor variations over the entity name are acceptable: (e.g. Amazon River vs River Amazon) 
\end{itemize}

\noindent\textbf{Examples of true claims:}
\begin{itemize}
\item Keanu Reeves has acted in a Shakespeare play
\item The Assassin’s Creed game franchise was launched in 2007
\item Prince Hamlet is the Prince of Denmark
\item In 2004, the coach of the Argentinian men’s national field hockey team was Carlos Retegui 
\end{itemize}

\subsection{Task 1 (subtask 1) Guidelines}
The objective of this task is to generate true claims from this source sentence that was extracted from Wikipedia.
\begin{itemize}
\item Extract simple factoid claims about {{entity}} given the source sentence. 
\item Use the source sentence and dictionary as the basis for your claims.
\item Reference any entity directly (i.e. pronouns and nominals should not be used)
\item Minor variations of names are acceptable (e.g. John F Kennedy, JFK, President Kennedy). 
\item Avoid vague or cautions language (e.g. might be, may be, could be, is reported that)
\item Correct capitalisation of entity names should be followed (India, not india).
\item Sentences should end with a period.
\item Numbers can be formatted in any appropriate English format (including as words for smaller quantities).
\item Some of the extracted text might not be accurate. These are still valid candidates for summary. It is not your job to fact check the information
\end{itemize}

\noindent\textbf{World Knowledge}
\begin{itemize}
\item Do not incorporate your own knowledge or beliefs. 
\item Additional world knowledge is given to the you in the form of a dictionary. Use this to make more complex claims 
(we prefer using this dictionary instead of your own knowledge because the information in this dictionary can be backed up from Wikipedia)
\item If the source sentence is not suitable, leave the box blank to skip.
\item If a dictionary entry is not suitable or uninformative, ignore it.
\end{itemize}

\begin{figure*}
\centering
\includegraphics[width=.6\textwidth]{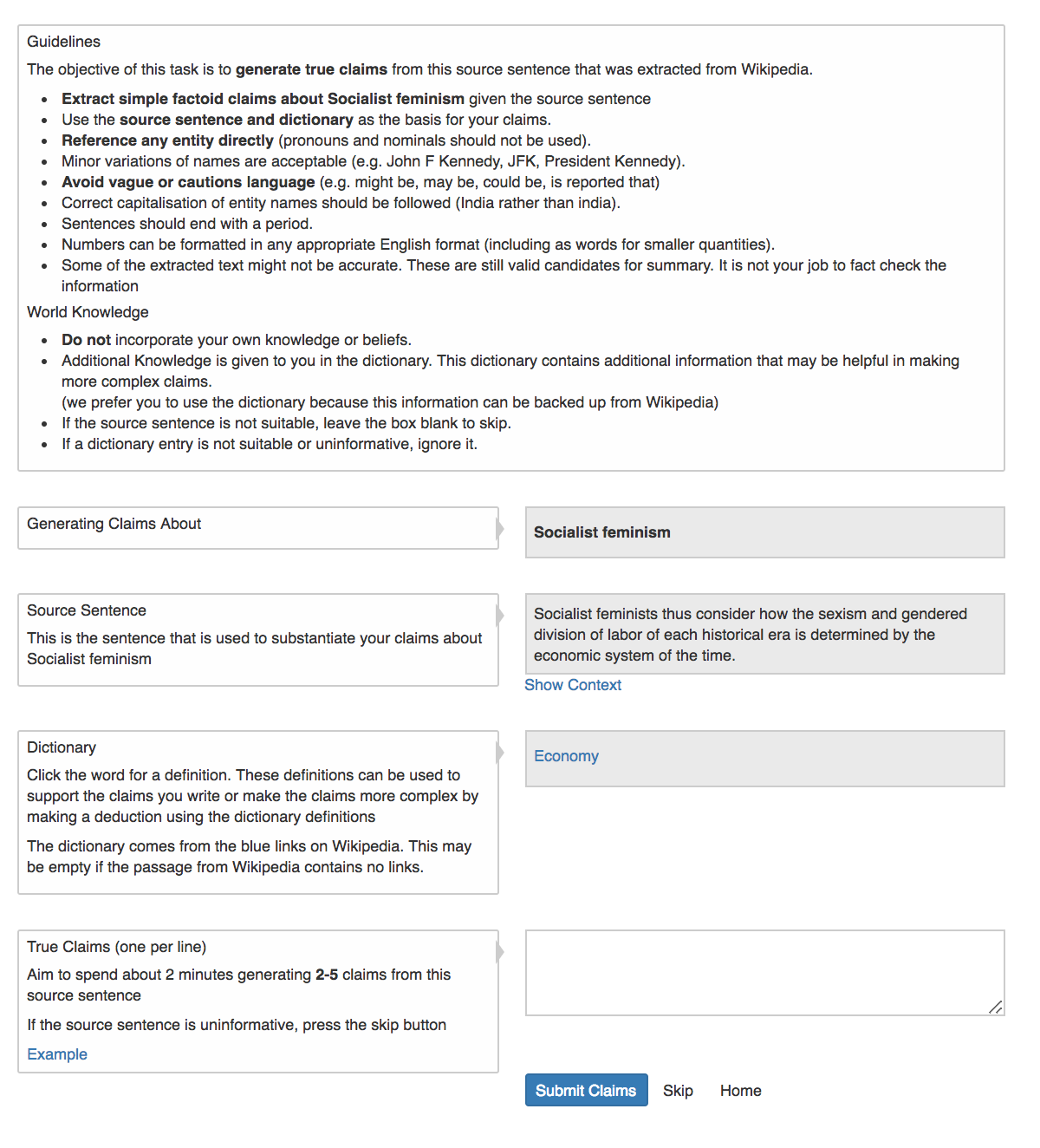}
\caption{Screenshot of annotation task 1 (subtask 1)}
\end{figure*}

\subsection{Task 1 (subtask 1) Examples}
See Tables \ref{tbl:task1ex-india} and \ref{tbl:task1ex-canada} for examples from the real data.
\begin{table*}[tbp]
\centering
\begin{tabular}{ll}
\textbf{Entity}  & \textsc{India} \\
\hline
\textbf{Source Sentence} & \begin{tabular}[c]{@{}l@{}}It shares land borders with Pakistan to the west; China, Nepal, and Bhutan \\ to the northeast; and Myanmar (Burma) and Bangladesh to the east. \end{tabular}  \\
\hline
\textbf{Dictionary} & \begin{tabular}[c]{@{}l@{}}\textbf{Bhutan}\\ Bhutan, officially the Kingdom of Bhutan, is a landlocked country in Asia, \\ and it is the smallest state located entirely within the Himalaya mountain range.\\ 
\textbf{China}\\ China, officially the People's Republic of China (PRC), is a unitary sovereign \\ state in East Asia and the world's most populous country, with a population of \\ over 1.381 billion.\\
\textbf{Pakistan} \\
Pakistan, officially the Islamic Republic of Pakistan, is a federal parliamentary \\ republic in South Asia on the crossroads of Central and Western Asia.
\end{tabular} \\
\hline
\textbf{Claims} & \begin{tabular}[c]{@{}l@{}}- One of the land borders that India shares is with the world's most \\populous country.\\ (uses information from the dictionary entry for china)\\ 
- India borders 6 countries.\\ (summarises some of the information in the source sentence)\\ 
- The Republic of India is situated between Pakistan and Burma.\\ (deduced by Pakistan being West of India, and Burma being to the East)\\\end{tabular}
\end{tabular}
\caption{Task 1 (subtask 1) example: India}
\label{tbl:task1ex-india}
\end{table*}

\begin{table*}[tbp]
\centering
\begin{tabular}{ll}
\textbf{Entity}  & \textsc{Canada} \\
\hline
\textbf{Source Sentence} & \begin{tabular}[c]{@{}l@{}}Canada is sparsely populated, the majority of its land territory being \\dominated by forest and tundra and the Rocky Mountains. \end{tabular}  \\
\hline
\textbf{Dictionary} & \begin{tabular}[c]{@{}l@{}}\textbf{Province of Canada}\\ The United Province of Canada, or the Province of Canada, or the \\United Canadas was a British colony in North America from 1841 to 1867.\\ 
\textbf{Rocky Mountains}\\ The Rocky Mountains, commonly known as the Rockies, are a major \\mountain range in western North America.\\
\textbf{tundra} \\ In physical geography, tundra is a type of biome where the tree growth \\ is hindered by low temperatures and short growing season.
\end{tabular} \\
\hline
\textbf{Claims} & \begin{tabular}[c]{@{}l@{}}- The terrain in Canada is mostly forest and tundra.\\ 
- Parts of Canada are subject to low temperatures.\\ 
- Canada is in North America.\\
- In some areas of Canada, it is difficult for trees to grow.\end{tabular}
\end{tabular}
\caption{Task 1 (subtask 1) example: Canada}
\label{tbl:task1ex-canada}
\end{table*}

\subsection{Task 1 (substask 2) Guidelines}
The objective of this task is to generate modifications to claims. The modifications can be either true or false. You will be given specific instructions about the types of modifications to make.

\begin{figure*}
    \centering
    \begin{subfigure}[b]{0.45\textwidth}
        \includegraphics[width=\textwidth]{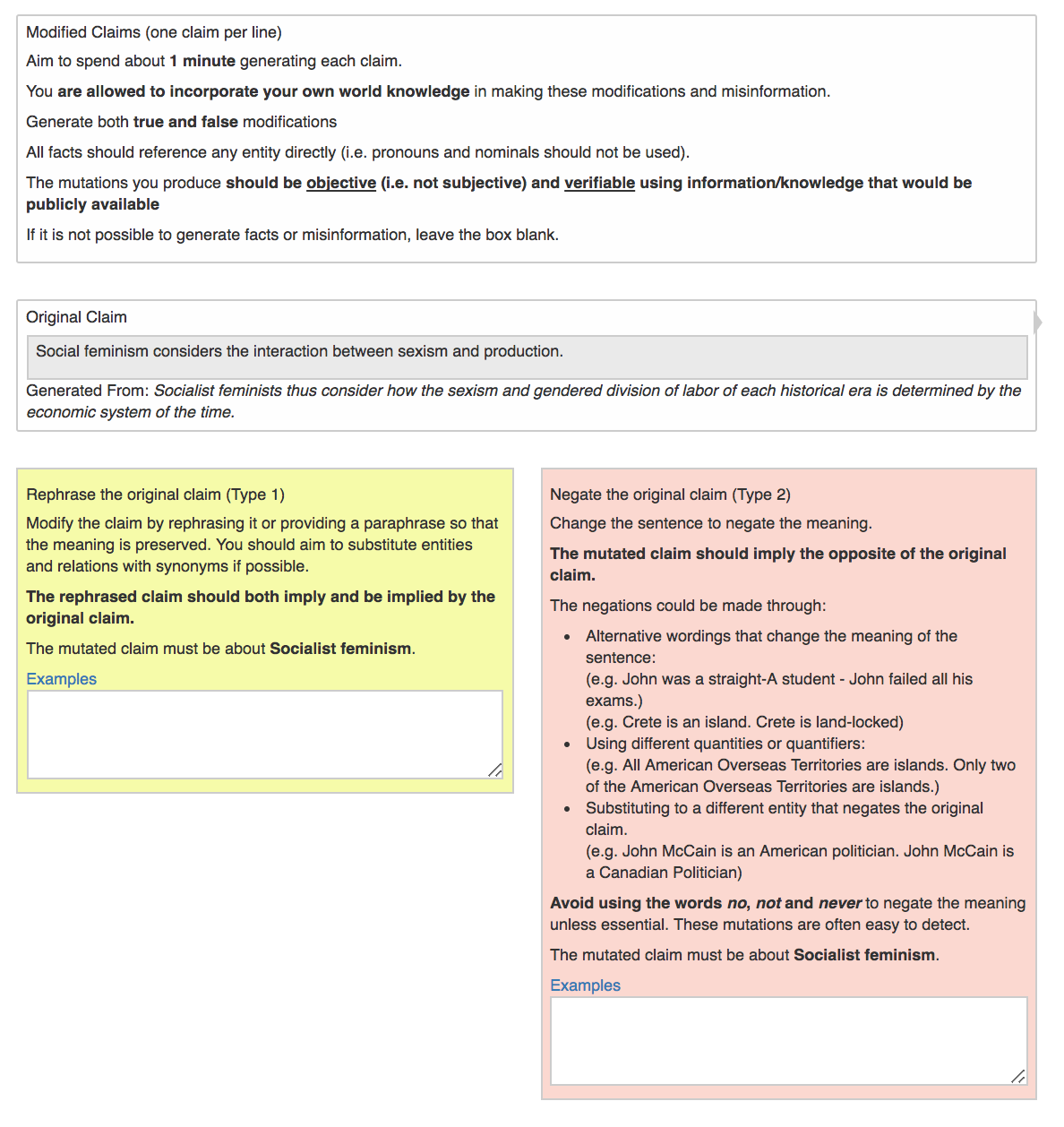}
        \caption{}
    \end{subfigure}
    ~ %add desired spacing between images, e. g. ~, \quad, \qquad, \hfill etc. 
    \begin{subfigure}[b]{0.45\textwidth}
        \includegraphics[width=\textwidth]{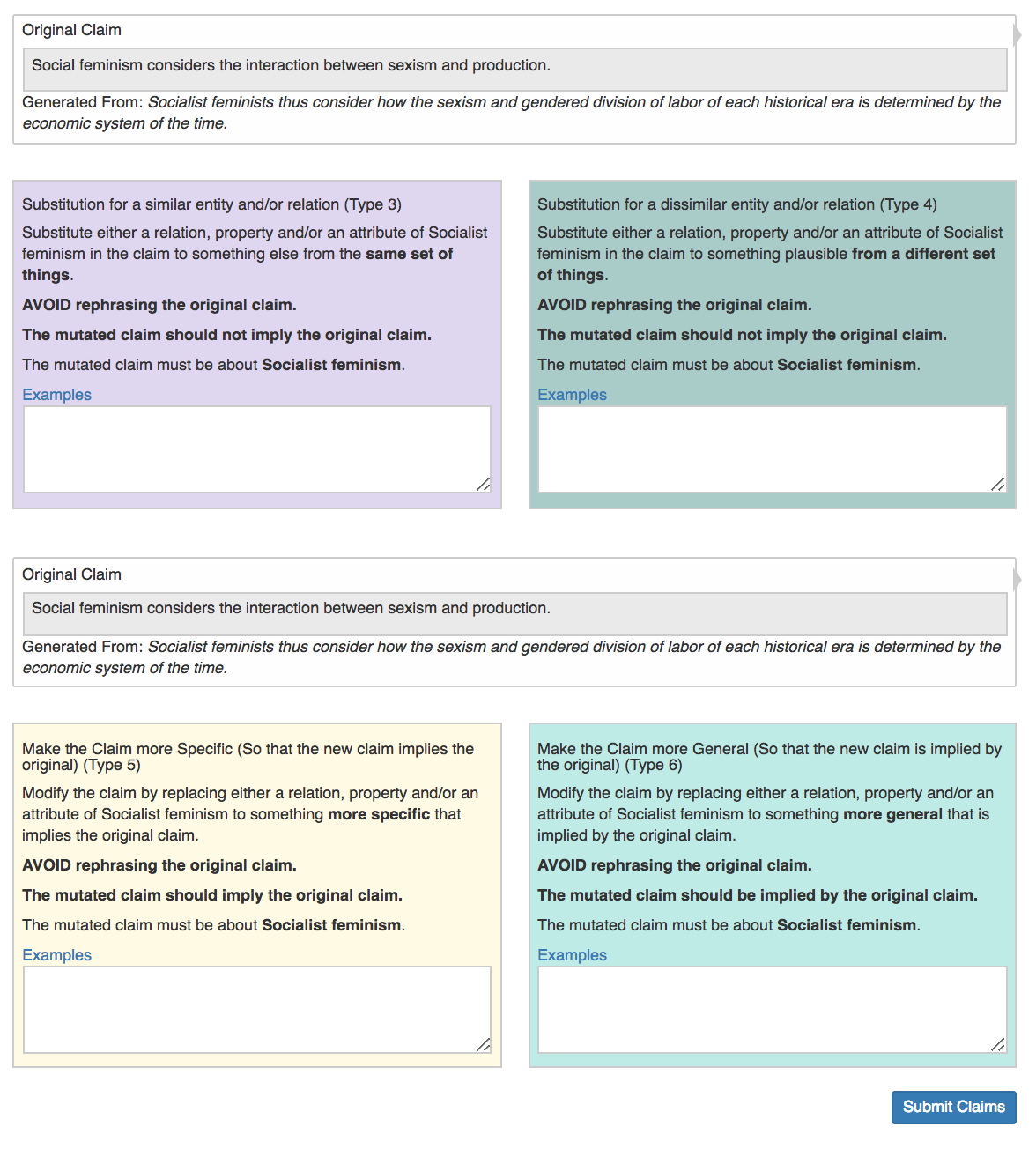}
        \caption{}
    \end{subfigure}
    \caption{Screenshots of annotation task 1 (subtask 2)}
    \label{fig:task1-2}
\end{figure*}

\begin{itemize}
\item Use the original claims and dictionary as the basis for your modifications to facts about {{entity}}
\item Reference any entity directly (i.e. pronouns and nominals should not be used)
\item Avoid vague or cautions language (e.g. might be, may be, could be, is reported that)
\item Correct capitalisation of entity names should be followed (India, not india).
\item Sentences should end with a period.
\item Numbers can be formatted in any appropriate English format (including as words for smaller quantities).
\item Some of the extracted text might not be accurate. These are still valid candidates for summary. It is not your job to fact check the information
\end{itemize}

Specific guidelines for this screen

\begin{itemize}
\item Aim to spend about up to 1 minute generating each claim.
\item You are allowed to incorporate your own world knowledge in making these modifications and misinformation.
\item All facts should reference any entity directly (i.e. pronouns and nominals should not be used).
\item The mutations you produce should be objective (i.e. not subjective) and verifiable using information/knowledge that would be publicly available
\item If it is not possible to generate facts or misinformation, leave the box blank.
\end{itemize}

There are six types of mutation the annotator will be asked to introduce. These will all be given on the same annotation page as all the claim modification types are related. 

\begin{enumerate}
\item Rephrase the claim so that it has the same meaning
\item Negate the meaning of the claim
\item Substitute the verb and/or object in the claim to alternative from the same set of things
\item Substitute the verb and/or object in the claim to alternative from a different set of things
\item Make the claim more specific so that the new claim implies the original claim (by making the meaning more specific)
\item Make the claim more general so that the new claim can be implied by the original claim (by making the meaning less specific)
\end{enumerate}

It may not always be possible to generate claims for each modification type. In this case, the box may be left blank.

\subsection{Task 1 (subtask 2) Examples}
\begin{figure}
\centering
\includegraphics[width=.6\columnwidth]{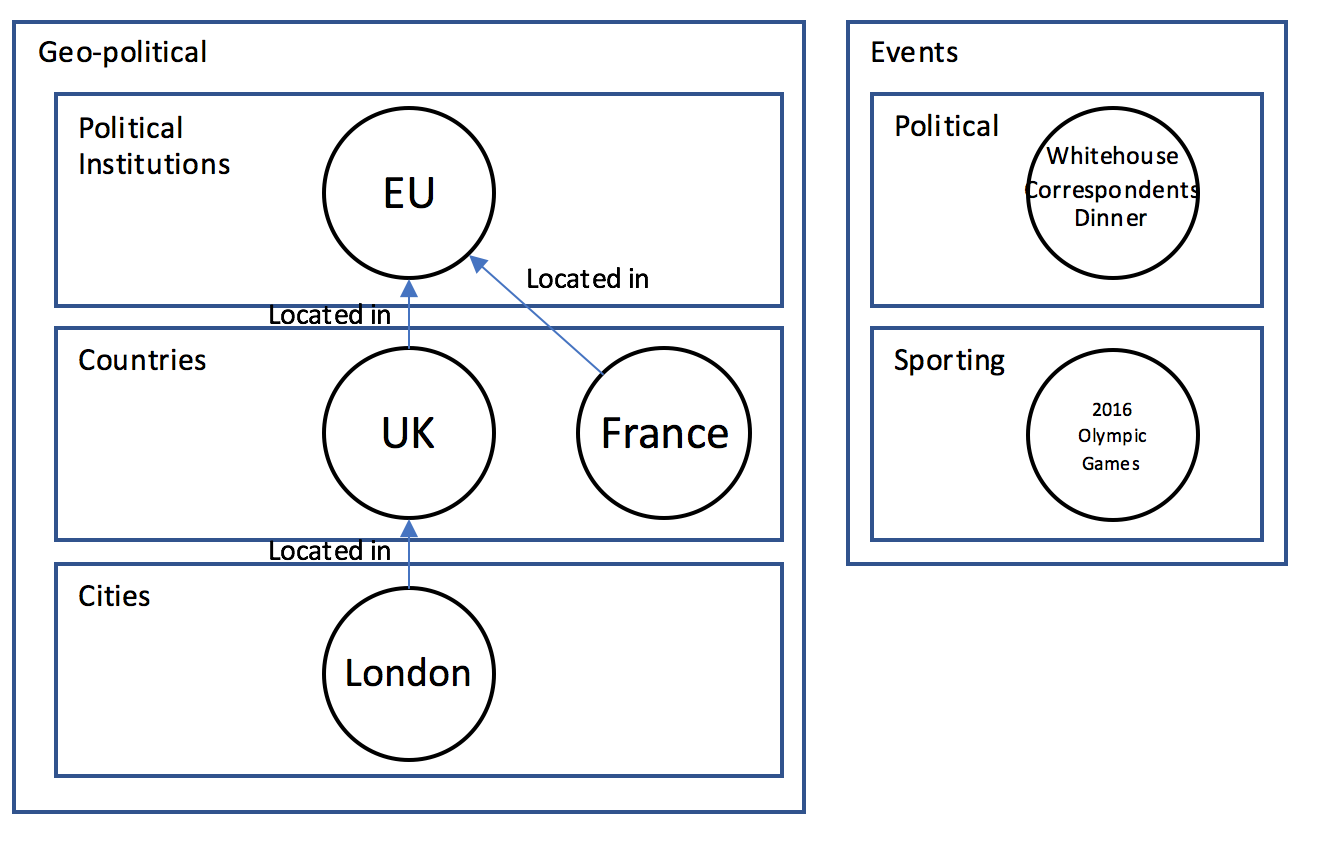}
\caption{Toy ontology to be used with the provided examples of similar and dissimilar mutations}
\label{fig:task1-ontology}
\end{figure}

The following example illustrates how given a single source sentence, the following mutations could be made and why they are suitable. For the claim ``\textit{Barack Obama toured the UK.}'', 
Figure \ref{fig:task1-ontology} shows the relations between objects, and Table \ref{tbl:task1ex-obama} contains examples for each type of mutation.
\begin{table*}[tbp]
\centering
\bgroup
\def\arraystretch{1.5}
\centering
\begin{tabular}{lp{5cm}p{5cm}}
\textbf{Type}             & \textbf{Claim}                                               & \textbf{Rationale}                                                                         \\ \hline
Rephrase              &  President Obama visited some places in the United Kingdom. & Rephrased. Same meaning.                                                                   \\
Negate                & Obama has never been to the UK before.                       & Obama could not have toured the UK if he has never been there.                             \\
Substitute Similar    & Barack Obama visited France.                                 & Both the UK and France are countries                                                       \\
Substitute Dissimilar & Barrack Obama attended the Whitehouse Correspondents Dinner.& In the claim, Barack Obama is visiting a country, whereas the dinner is a political event. \\
More specific         & Barrack Obama made state visit to London.                    & London is in the UK. If Obama visited London, he must have visited the UK.                 \\
More general          & Barrack Obama visited a country in the EU.                   & The UK is in the EU. If Obama visited the UK, he visited an EU country.                 
\end{tabular}
\egroup
\caption{Example mutations}
\label{tbl:task1ex-obama}
\end{table*}

\subsection{Task 2 Guidelines}
The purpose of this task is to identify evidence from a Wikipedia page that can be used to support or refute simple factoid sentences called claims. The claims are generated by humans (as part of the WF1 annotation workflow) from a Wikipedia page. Some claims are true. Some claims are fake. You must find the evidence from the page that supports or refutes the claim.

Other Wikipedia pages will also provide additional information that can serves as evidence. For each line, we will provide extracts from the linked pages in the dictionary column which appear when you ``Expand'' the sentence. The sentences from these linked pages that contain relevant supplementary information should be individually selected to record which information is used in justifying your decisions.

Step-by-step guide:
\begin{enumerate}
\item Read and understand the claim
\item Read the Wikipedia page and identify sentences that contain relevant information.
\item On identifying a relevant sentence, press the Expand button to highlight it. This will load the dictionary and the buttons to annotate it:
\begin{enumerate}
\item If the highlighted sentence contains enough information in a definitive statement to support or refute the claim, press the Supports or Refutes button to add your annotation. No information from the dictionary is needed in this case (this includes information from the main Wikipedia page). Then continue annotating from step 2.
\item If the highlighted sentence contains some information supporting or refuting the claim but also needs supporting information, this can be added from the dictionary. 
\begin{enumerate}
\item The hyperlinked sentences from the passage are automatically added to the dictionary
\item If a sentence from the main Wikipedia article is needed to provide supporting information. Click “Add Main Wikipedia Page” to add it to the dictionary. 
\item If the claim or sentence contains an entity that is not in the dictionary, then a custom page can be added by clicking “Add Custom Page”. Use a search engine of your choice to find the page and then paste the Wikipedia URL into the box.
\item Tick the sentences from the dictionary that provide the minimal amount of supporting information needed to form your decision. If there are multiple equally relevant entries (such as a list of movies), then just select the first.
Once all required information is added, then press the Supports or Refutes button to add your annotation and continue from step 2.
\end{enumerate}
\item If the highlighted sentence and the dictionary do not contain enough information to support or refute the claim, press the Cancel button and continue from step 2 to identify more relevant sentences.
\end{enumerate}
\item On reaching the end of the Wikipedia page. Press Submit if you could find information that supports or refutes the claim. If you could not find any supporting evidence, press Skip then select Not enough information
\end{enumerate}

The objective is to find sentences that support or refute the claim.

You must apply common-sense reasoning to the evidence you read but avoid applying your own world-knowledge by basing your decisions on the information presented in the Wikipedia page and dictionary.

As a guide - you should ask yourself:

If I was given only the selected sentences, do I have stronger reason to believe claim is true (supported) or stronger reason to believe the claim is false (refuted). If I'm not certain, what additional information (dictionary) do I have to add to reach this conclusion.

\begin{figure*}
\centering
\includegraphics[width=.6\textwidth]{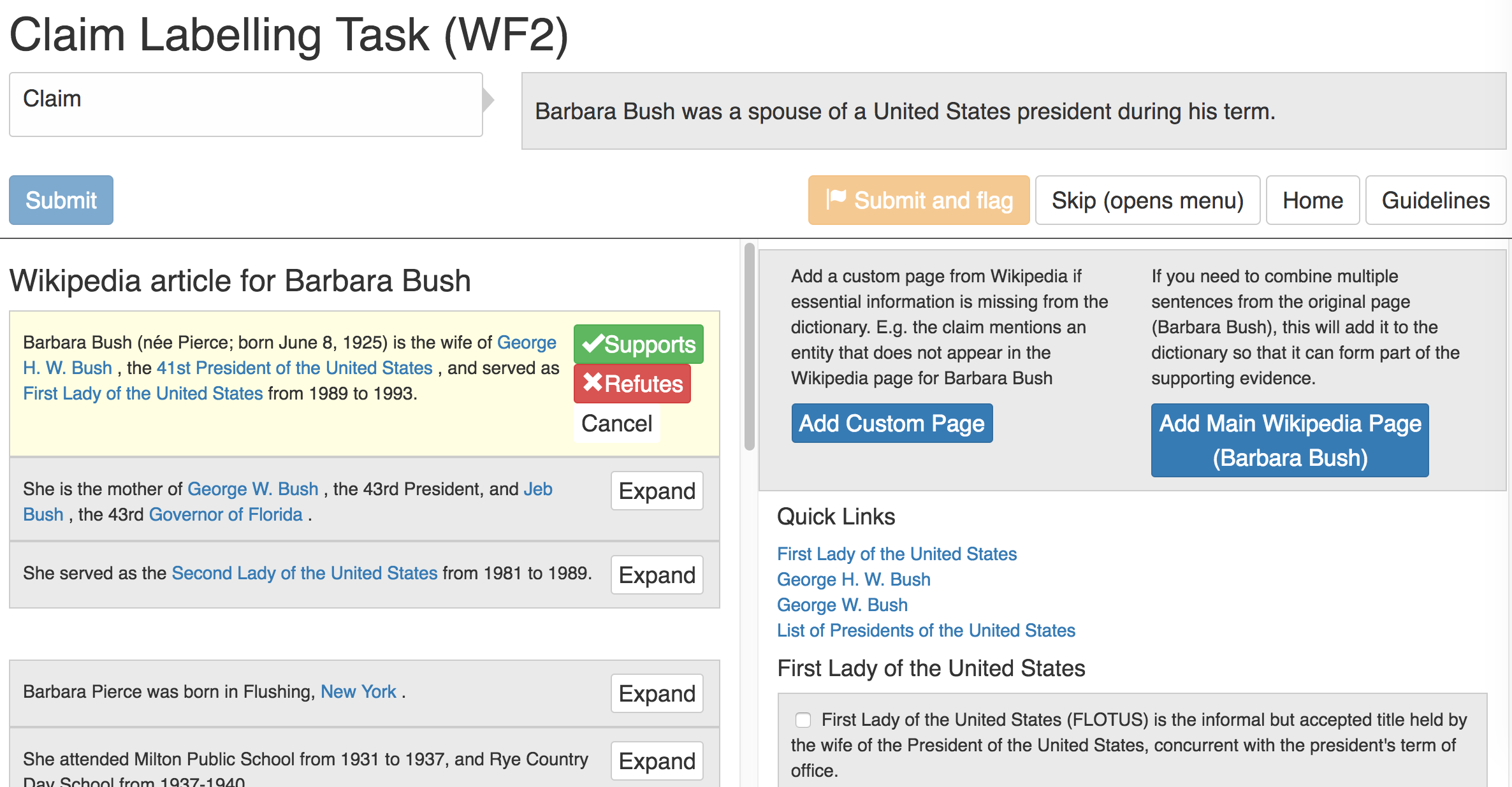}
\caption{Screenshot of annotation task 2}
\end{figure*}

\subsection{Task 2 Examples}
\subsubsection{What does it mean to Support or Refute a claim}

The following count as valid justifications for marking an item as supported/refuted:

Sentence directly states information that supports/refutes the claim or states information that is synonymous/antonymous with information in the claim

Claim: Water occurs artificially\\
Refuted by: ``It also occurs in nature as snow, glaciers ...''

Claim: Samuel L. Jackson was in the third movie in the Die Hard film series.\\
Supported by: ``He is a highly prolific actor, having appeared in over 100 films, including Die Hard 3.''

Sentence refutes the claim through negation or quantification

Claim: Schindler's List received no awards.\\
Refuted by: ``It was the recipient of seven Academy Awards (out of twelve nominations), including Best Picture, Best Director...''

Sentence provides information about a different entity and only one entity is permitted (e.g. place of birth can only be one place)

Claim: David Schwimmer finished acting in Friends in 2005.\\
Refuted by: ``After the series finale of Friends in 2004, Schwimmer was cast as the title character in the 2005 drama Duane Hopwood.''

Sentence provides information that, in conjunction with other sentences from the dictionary, fulfils one of the above criteria

Claim: John McCain is a conservative.\\
Refuted by: ``He was the Republican nominee for the 2008 U.S. presidential election.'' AND ``The Republican Party's current ideology is American conservatism, which contrasts with the Democrats' more progressive platform (also called modern liberalism).''

\paragraph{Adding Main Wikipedia Page to Dictionary}
In the case where the claim can be supported from multiple sentences from the main Wikipedia page, information the main Wikipedia page should be added to the dictionary to add supporting information.  This is because each line that is annotated in the left column for the main Wikipedia page is stored independently.

\begin{description}
\item[Claim:] George Washington was a soldier, born in 1732.

\item[Wikipedia page:] George Washington

Sentence 1: George Washington was an American politician and soldier who served as the first President of the United States from 1789 to 1797 and was one of the Founding Fathers of the United States.

Sentence 2: In 1775, the Second Continental Congress commissioned him as commander-in-chief of the Continental Army in the American Revolution.

Sentence 3: The Gregorian calendar was adopted within the British Empire in 1752, and it renders a birth date of February 22, 1732.
\end{description}
Sentence 1 contains enough information to wholly support the claim without the need for any additional information.

Sentence 2 and 3 contain partial information that can be combined. Expand sentence 2 and click add main Wikipedia page to add the Wikipedia page to add George Washington to the dictionary. Sentence 3 can now be added to dictionary to support the claim.

The order of the sentences doesn’t matter (selecting sentence 2+3 is the same as adding sentence 3+2) because we sort the sentences in document order. This means that you only need to annotate this once. 

If you attempt to add the main Wikipedia page to the dictionary from sentence 3 having already used it for sentence 2, the system will warn you that you are making a duplicate annotation.

\subsubsection{Adding Custom Pages}
You may need to add a custom page from Wikipedia to the dictionary. This may happen in cases where the claim discusses an entity that was not in the original Wikipedia page

Claim: Colin Firth is a Gemini.
In Original Page: ``Colin Firth (born 10 September 1960)... ''
Requires Additional Information from Gemini: ``Under the tropical zodiac, the sun transits this sign between May 21 and June 21.''

Tense
The difference in verb tenses that do not affect the meaning should be ignored.

Claim: Frank Sinatra is a musician
Supported: He is one of the best-selling music artists of all time, having sold more than 150 million records worldwide.

Claim: Frank Sinatra is a musician 
Supported: Francis Albert Sinatra was an American singer

\subsubsection{Skipping}
There may be times where it is appropriate to skip the claim by pressing the Skip button:

\begin{itemize}
\item The claim cannot be verified using the information with the information provided:
\begin{itemize}
  \item If the claim could potentially be verified using other publicly available information. Select Not Enough Information
  \item If the claim can't be verified using any publicly available information (because it's ambiguous, vague, personal or implausible) select The claim is ambiguous or contains personal information
  \item If the claim doesn’t meet the guidelines from WF1, select: The claim doesn’t meet the WF1 guidelines
\end{itemize}
\item The claim contains typographical errors, spelling mistakes, is ungrammatical or could be fixed with a very minor change
\begin{itemize}
  \item Select The claim has a typo or grammatical error
\end{itemize}
\end{itemize}

Keep in mind that clicking Not Enough Information or The claim is ambiguous or contains personal information is still very useful feedback for the AI systems. They need examples of what a verifiable claim looks like, and negative examples are as useful (if not more so) than positive ones!

\subsection{Task 2 additional guidelines}
After conferring with the organizers, the annotators expanded the guidelines to include common case that were not explicitly covered in the guidelines:

\begin{enumerate}
\item Any claims involving “many”, “several”, “rarely”, “barely” or other indeterminate count words are going to be ambiguous and can be flagged.
\item Same goes for “popular”, “famous”, and “successful (for people, for works we can assume commercial success)”
\item We cannot prove personhood for fictional characters like we can for real people (dogs and cats can be authors, actors, and citizens in fiction)
\item If a claim is “[Person] was in [film]”, the only way to refute it would be (a) if they were born after it was released, or (b) their acting debut is mentioned and occurs a realistically long enough amount of time after the film’s release (at least a few years).
\item A list of movies that someone was in or jobs a person held is not necessarily exclusive, we cannot refute someone being a lawyer because the first sentence of their wiki article says they were an actor.
\item A person is not their roles, if a claim is something like “Tom Cruise participated in a heist in Mission Impossible 3”, we cannot prove it, because Ethan Hunt did that, not Tom Cruise.
\item Our workflow is time-insensitive, but if a claim tags something to a time period, we can treat it as such. “Neil Armstrong is an astronaut” can be supported, but “Neil Armstrong was an astronaut in 2013” can be refuted, because he was dead at the time.
\item If someone won 5 Academy Awards, they won 3 Academy Awards. Similarly, if they won an Academy Award, they were nominated for an award.
\item Multiple citizenships can exist.
\item If a claim says “[Person] was in [film] in 2009”, then the film’s release date can support it. If the claim is “[Person] acted in [film] in 2009”, filming dates or release dates can prove it.
\item Flag anything related to death, large-scale recent disasters, and controversial religious or social statements.
\end{enumerate}

\section{Manual Error Analysis}
\label{appdx:manual}
\begin{figure*}
\centering
\includegraphics[width=\linewidth]{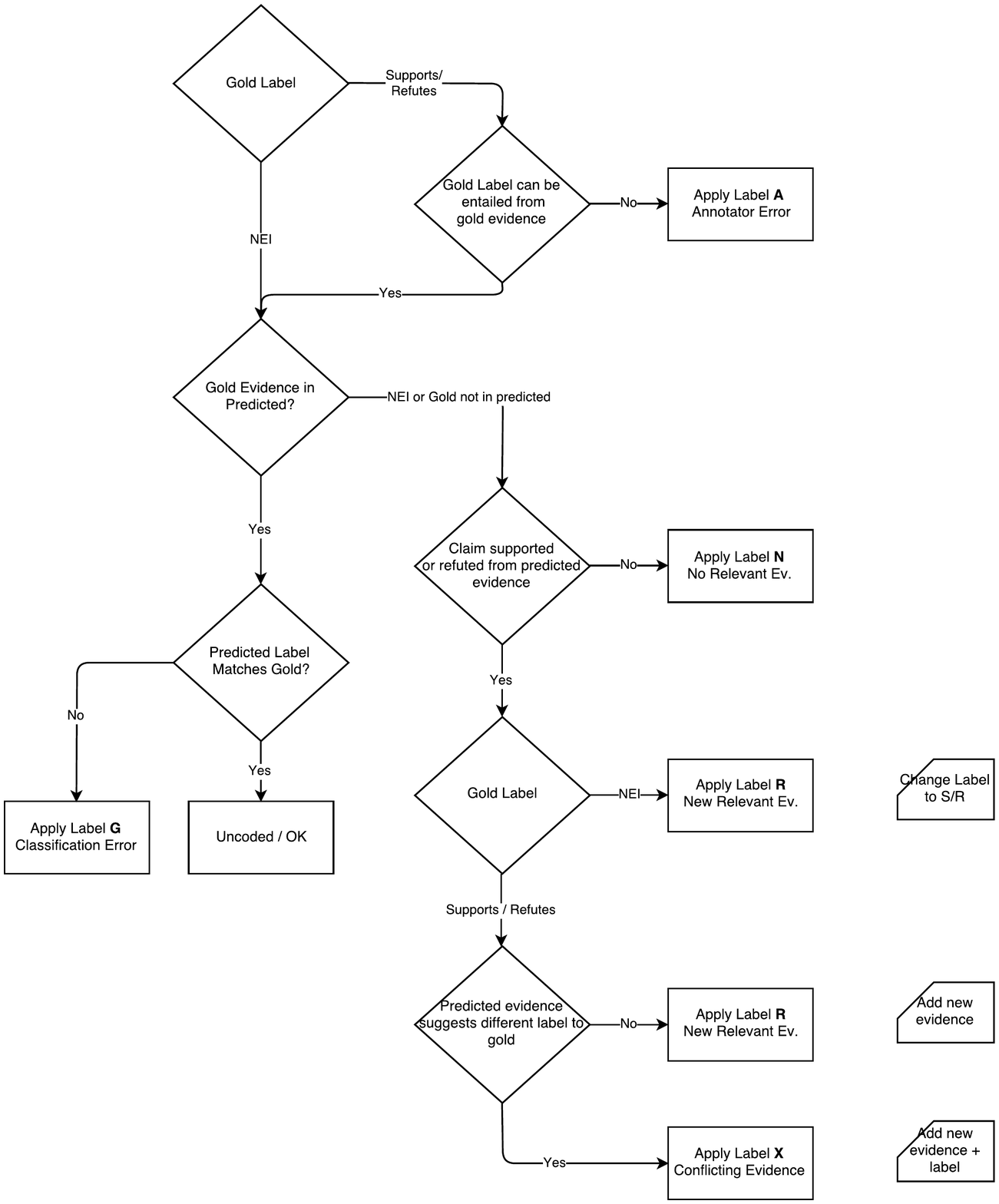}
\caption{Manual Error Coding Process}
\label{fig:error_analysis}
\end{figure*}
The manual error analysis was conducted with the  decision process described in Figure~\ref{fig:error_analysis}. For the cases of finding new evidence, recommended actions have been listed. While these were not performed for this version of the dataset, this may form a future update following a pool-based evaluation in the FEVER Shared Task.